\documentclass[11pt,a4paper]{article}

\PassOptionsToPackage{table}{xcolor}
\usepackage{polyu-vclab}
\usepackage[authoryear,round]{natbib}

\usepackage{amsmath,amsfonts,bm}

\def\eqref#1{equation~\ref{#1}}

\def\1{\bm{1}}

\DeclareMathAlphabet{\mathsfit}{\encodingdefault}{\sfdefault}{m}{sl}
\SetMathAlphabet{\mathsfit}{bold}{\encodingdefault}{\sfdefault}{bx}{n}

\usepackage[utf8]{inputenc}
\usepackage[T1]{fontenc}
\usepackage{url}
\usepackage{booktabs}
\usepackage{amsmath}
\usepackage{amsfonts}
\usepackage{nicefrac}
\usepackage{microtype}
\usepackage{xcolor}
\usepackage{graphicx}
\usepackage{multirow}
\usepackage{pifont}
\usepackage{colortbl}
\usepackage{threeparttable}
\usepackage{xspace}
\usepackage{todonotes}
\usepackage[etex=true,export]{adjustbox}
\usepackage{makecell}
\usepackage{subcaption}
\usepackage{epsfig}
\usepackage{epstopdf}
\usepackage{bbding}
\usepackage{diagbox}
\usepackage{footnote}
\usepackage{wrapfig}
\usepackage[normalem]{ulem}
\usepackage{enumitem}
\usepackage{tikz}
\usepackage{algorithm}
\usepackage{algpseudocode}
\usepackage{textcomp}
\usepackage{chngcntr}
\usepackage{array}

\definecolor{citecolor}{HTML}{0071bc}
\hypersetup{colorlinks=true,linkcolor=polyureddark,citecolor=citecolor}
\usepackage[capitalise]{cleveref}

\newcommand{\cmark}{\textcolor{green!60!black}{\ding{51}}}
\newcommand{\xmark}{\textcolor{red}{\ding{55}}}

\definecolor{omniblue}{HTML}{EEF5FB}
\definecolor{omnigreen}{HTML}{F2F8EC}
\definecolor{omniyellow}{HTML}{FFFCE8}
\definecolor{promptblue}{HTML}{243B55}
\definecolor{promptline}{HTML}{B7C4D3}
\definecolor{promptfill}{HTML}{F7F9FC}
\definecolor{promptsectionfill}{HTML}{E9EEF5}
\usepackage[most]{tcolorbox}

\newtcolorbox{omniprompt}[2][]{
  enhanced,
  title={#2},
  colback=white,
  colframe=black!70,
  colbacktitle=black!70,
  coltitle=white,
  fonttitle=\bfseries\small,
  arc=4pt,
  boxrule=0.7pt,
  left=8pt,
  right=8pt,
  top=4pt,
  bottom=4pt,
  before skip=2pt,
  after skip=2pt,
  #1
}

\newcommand{\evalsectionlabel}[1]{%
  \par\vspace{2pt}
  {\normalfont\bfseries #1\par}
  \vspace{3pt}
}

\newcommand{\evalinputpromptdivider}{%
  \par\vspace{3pt}
  \noindent
  \begin{tikzpicture}
    \useasboundingbox (0,-0.5pt) rectangle (\linewidth,0.5pt);
    \draw[densely dashed,draw=black!38,line width=0.55pt]
      (0,0) -- (\linewidth,0);
  \end{tikzpicture}
  \par\vspace{1pt}
}

\newcommand{\evalvideoinput}{%
  \begin{minipage}[c][2.45cm][c]{\linewidth}
    \centering
    \includegraphics[
      width=0.88\linewidth,
      height=1.72cm,
      keepaspectratio
    ]{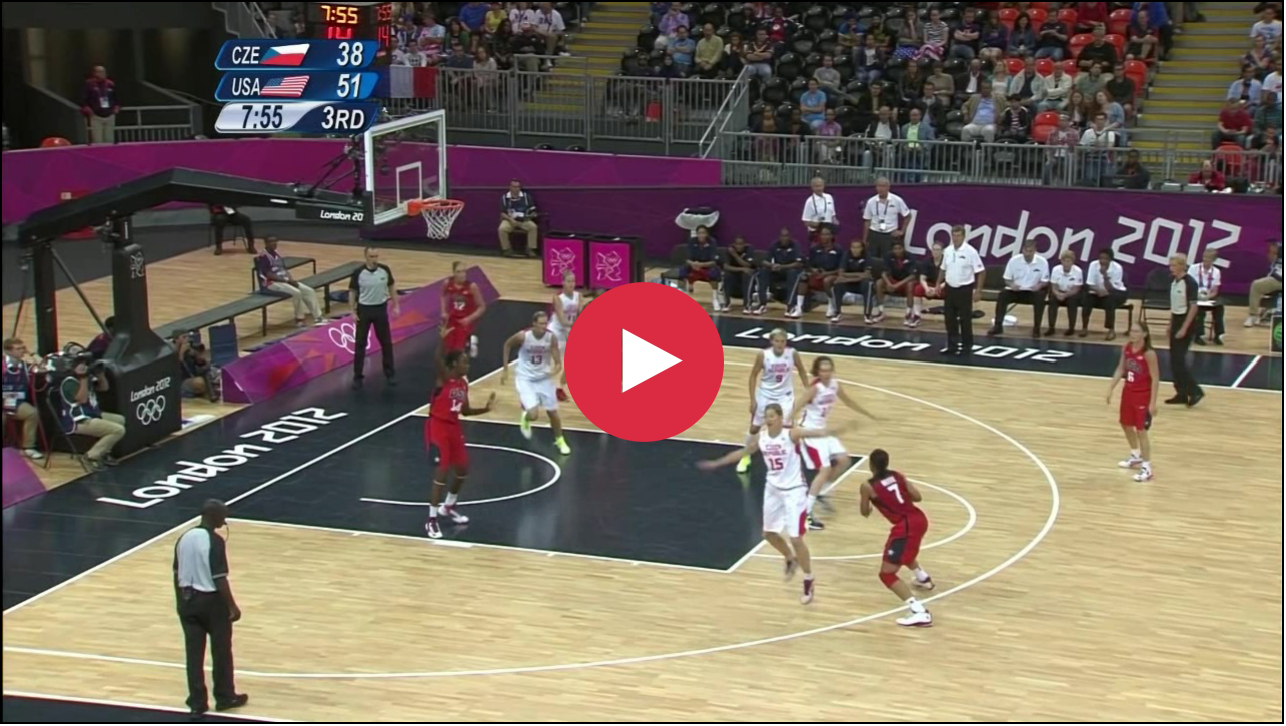}\\[-1pt]
    {\scriptsize Video frames}
  \end{minipage}
}

\newcommand{\evalreferencevideoinput}[1]{%
  \begin{minipage}[c][2.45cm][c]{\linewidth}
    \centering
    \begin{tabular}{@{}c@{\hspace{4pt}}c@{\hspace{4pt}}c@{}}
      \includegraphics[
        height=1.72cm,
        width=0.56\linewidth,
        keepaspectratio
      ]{Figures/prompt_presentation/video.png}
      &
      {\large$+$}
      &
      #1
      \\
      {\scriptsize Video frames} & & {\scriptsize Reference image}
    \end{tabular}
  \end{minipage}
}

\newcommand{\evalvideoreferenceinput}[1]{\evalreferencevideoinput{#1}}

\newcommand{\evalavvideoinput}{%
  \begin{minipage}[c][2.45cm][c]{\linewidth}
    \centering
    \includegraphics[
      width=0.88\linewidth,
      height=1.72cm,
      keepaspectratio
    ]{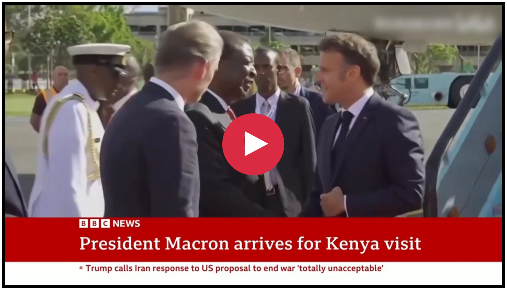}\\[-1pt]
    {\scriptsize Video frames with audio}
  \end{minipage}
}

\newcommand{\evalavreferencevideoinput}[1]{%
  \begin{minipage}[c][2.45cm][c]{\linewidth}
    \centering
    \begin{tabular}{@{}c@{\hspace{4pt}}c@{\hspace{4pt}}c@{}}
      \includegraphics[
        height=1.72cm,
        width=0.56\linewidth,
        keepaspectratio
      ]{Figures/prompt_presentation/video_av.png}
      &
      {\large$+$}
      &
      #1
      \\
      {\scriptsize Video frames with audio} & &
      {\scriptsize Reference image}
    \end{tabular}
  \end{minipage}
}

\newcommand{\evalpromptoutput}[1]{%
  \tcblower
  {\normalfont\bfseries Output\par}
  \smallskip
  \strut\textbf{Ground Truth:} #1%
}

\newenvironment{evalpromptbody}[1]
  {\begin{minipage}[t][#1][t]{\linewidth}}
  {\end{minipage}}

\newtcolorbox{evalpromptcard}[2][]{
  enhanced,
  height=#2,
  colback=white,
  colframe=black!70,
  colbacktitle=black!70,
  coltitle=white,
  fonttitle=\bfseries\small,
  title=#1,
  boxrule=0.7pt,
  arc=4pt,
  left=8pt,
  right=8pt,
  top=4pt,
  bottom=4pt,
  toptitle=4pt,
  bottomtitle=4pt,
  before skip=2pt,
  after skip=2pt
}

\newcommand{\evalpromptintrovideo}{%
  You are given a sequence of frames uniformly sampled across a 52-second video.
  Each video frame is preceded by its timestamp in seconds.
  Complete the task based on the visual content of these video frames.
}

\newcommand{\evalpromptintroref}{%
  You are given a sequence of frames uniformly sampled across a 52-second video,
  followed by a reference image. Each video frame is preceded by its timestamp in seconds. Complete the task based on the visual content of the video frames
  and the reference image.
}

\newcommand{\evalpromptintroavvideo}{%
  You are given a sequence of frames uniformly sampled across a 270-second video and the corresponding audio.
  Each frame is preceded by its timestamp in seconds.
  Complete the task using the video frames and audio.
}

\newcommand{\evalpromptintroavref}{%
  You are given a sequence of frames uniformly sampled across a 270-second video and the corresponding audio, followed by a reference image.
  Each frame is preceded by its timestamp in seconds.
  Complete the task using the video frames, audio, and reference image.
}

\setEyebrow{PolyU VCLab\,\textbullet\,Preprint 2026}

\papertitle{OmniCompass: A Holistic Benchmark for Omni-Modal Video Grounding and Understanding}

\paperauthors{%
  \raggedright
  \mbox{Xuyang Liu\affilmark{1,2}}\quad
  \mbox{Shihao Wang\affilmark{1}}\quad
  \mbox{Xinyu Wei\affilmark{1,2}}\quad
  \mbox{Minghan Li\affilmark{3}}\quad
  \mbox{Zeqing Wang\affilmark{1,2}}
  \\[3pt]
  \mbox{Zuhao Yang\affilmark{4}}\quad
  \mbox{Zichen Wen\affilmark{5}}\quad
  \mbox{Lei Zhang\affilmark{1,2}$^{\text{\Envelope}}$}
}

\paperaffil{%
  \raggedright
  \mbox{\affilmark{1}\,The Hong Kong Polytechnic University}\quad
  \mbox{\affilmark{2}\,OPPO Research Institute}\quad
  \mbox{\affilmark{3}\,Colorado School of Mines}\\[2pt]
  \mbox{\affilmark{4}\,Nanyang Technological University}\quad
  \mbox{\affilmark{5}\,Shanghai Jiao Tong University}%
}

\papernotes{}

\newcommand{\OmniCompassProjectURL}{https://omnicompass.netlify.app/}
\newcommand{\OmniCompassCodeURL}{https://github.com/PolyU-VCLab/OmniCompass}
\newcommand{\OmniCompassDatasetURL}{https://huggingface.co/datasets/VCLab-PolyU/OmniCompass}
\newcommand{\omnicompassbadge}[2]{%
  \begingroup
  \ifdefempty{#2}{\renewcommand{\href}[2]{##2}}{}%
  \vclabbadgesolid{#1}{#2}%
  \endgroup
}
\paperbadges{%
  \raggedright
  \omnicompassbadge{Homepage}{\OmniCompassProjectURL}\;%
  \omnicompassbadge{GitHub}{\OmniCompassCodeURL}\;%
  \omnicompassbadge{Hugging Face}{\OmniCompassDatasetURL}%
}

\renewenvironment{abstract}
  {\begin{vclabAbstract}\noindent\textbf{Abstract.}\;}
  {\end{vclabAbstract}}

\begin{document}

\maketitleVCLab
\begingroup
\makeatletter
\renewcommand{\thefootnote}{\Envelope}
\renewcommand{\@makefntext}[1]{%
  \noindent\@makefnmark\,#1%
}
\footnotetext[0]{%
  \begin{tabular}[t]{@{}l@{}}
  Corresponding author (cslzhang@comp.polyu.edu.hk).\\
  This work is supported by the PolyU-OPPO Joint Innovative Research Center.
  \end{tabular}%
}
\makeatother
\endgroup

\begin{abstract}
With the rapid development of multimodal large language models (MLLMs), evaluating their video understanding capabilities has become increasingly important.
Existing video understanding benchmarks cover only part of the evaluation dimensions, limiting systematic diagnosis of omni-modal video understanding.
We introduce \textbf{OmniCompass}, a holistic benchmark for omni-modal video grounding and understanding.
OmniCompass defines \textbf{two evaluation tracks}: \textbf{OmniCompass-VL} assesses vision-language models on tasks requiring visual evidence, while \textbf{OmniCompass-AV} assesses omni-modal models on tasks requiring joint audio-visual evidence.
Each track includes \textbf{three complementary tasks}: temporal grounding (TG), spatial grounding (SG), and question answering (QA), assessing \textit{when} a referenced target appears, \textit{where} it is, and \textit{what} happens.
QA covers five perception and three reasoning categories in OmniCompass-VL, and three perception and two reasoning categories in OmniCompass-AV.
OmniCompass-VL contains 820 videos and 2,460 instances across domains and \textbf{three duration types}: short, medium, and long.
A narrative- and audio-rich subset forms OmniCompass-AV with 240 videos and 720 instances.
Each instance is evaluated under \textbf{two paired settings}: text-only and text+image, specifying the same target through text and visual-language guidance.
All instances are \textbf{manually annotated} with multi-round quality verification.
Evaluations of \textbf{50} MLLMs show that open-source models still lag behind proprietary counterparts, particularly in audio-visual understanding.
Precise temporal and spatial grounding remains challenging even for leading models, and their predictions still exhibit inconsistencies across the paired target specifications.
\end{abstract}

\begin{figure}[h]
    \centering
    \includegraphics[width=\linewidth]{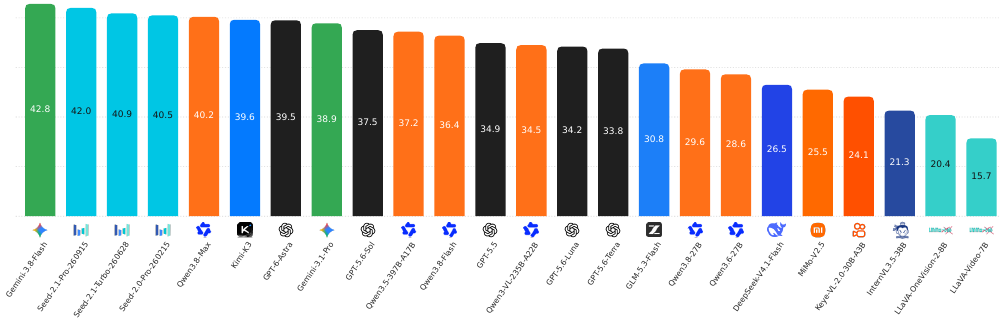}
    \vspace{-5mm}
    \caption{\textbf{Overall model performance on OmniCompass-VL.}
    Gemini-3.8-Flash and Kimi-K3 achieve the highest scores among proprietary and open-source models, respectively, at 42.8\% and 39.6\%.}
    \vspace{-5mm}
    \label{fig:overall_performance_vl}
\end{figure}
\section{Introduction}

Human perception uses linguistic or visual cues to identify targets, attend to visual and auditory evidence across space and time, and form goal-directed judgments.
Multimodal large language models (MLLMs) jointly process images, videos, audio, and text, supporting multimodal understanding and reasoning~\citep{Qwen3-VL,Qwen3-Omni,google2025gemini3}.
Recent MLLMs can revisit relevant temporal clips~\citep{yang2026longvt,zeng2026video-o3}, inspect task-relevant spatial evidence~\citep{wu2026vtoolr1,meng2025open-o3-video}, and selectively acquire audio-visual evidence during reasoning~\citep{xing2026omniagent,chen2026omnireasoner}.
These advances motivate evaluating omni-modal video grounding and understanding.
This requires identifying targets from textual or visual cues, grounding them spatiotemporally, and integrating audio-visual evidence to understand events.

\begin{table*}[t]
    \caption{
    \textbf{Comparison of existing video understanding benchmarks.}
    Target Specification indicates whether the target is specified using text alone (\textit{i.e.}, Text) or text with an image (\textit{i.e.}, Text+Image); Video Evidence indicates whether correct outputs depend on visual or native audio content. TG/SG/QA denote temporal grounding, spatial grounding, and question answering. Short/Med./Long denote video length of $<90$s, $[90,300)$s, and $\geq300$s. \#Vid./\#Inst. are video/instance counts; M/A/H denote manual/automatic/hybrid annotation; Len. is the average video length in seconds.
    }
    \vspace{-2mm}
    \label{tab:benchmark_comparison}
    \centering
    \setlength{\tabcolsep}{3.2pt}
    \renewcommand{\arraystretch}{1.04}
    \resizebox{\textwidth}{!}{
    \begin{threeparttable}
    \begin{tabular}{lcccccccccccccc}
    \toprule
    \multirow{2}{*}{\textbf{Benchmarks}}
    & \multicolumn{2}{c}{\textbf{Target Specification}}
    & \multicolumn{2}{c}{\textbf{Video Evidence}}
    & \multicolumn{3}{c}{\textbf{Task Coverage}}
    & \multicolumn{3}{c}{\textbf{Video Duration}}
    & \multirow{2}{*}{\textbf{\#Vid.}}
    & \multirow{2}{*}{\textbf{\#Inst.}}
    & \multirow{2}{*}{\textbf{Anno.}}
    & \multirow{2}{*}{\textbf{Len.(s)}} \\
    \cmidrule(lr){2-3} \cmidrule(lr){4-5} \cmidrule(lr){6-8} \cmidrule(lr){9-11}
    & \textbf{Text} & \textbf{Text+Image}
    & \textbf{Visual} & \textbf{Audio}
    & \textbf{TG} & \textbf{SG} & \textbf{QA}
    & \textbf{Short} & \textbf{Med.} & \textbf{Long}
    & & & & \\
    \midrule

    \rowcolor{omniblue}
    \multicolumn{15}{l}{\textbf{\textit{Video Benchmarks}}} \\
    \midrule

    Charades-STA~\citep{gao2017Charades-STA}
    & \cmark & \xmark & \cmark & \xmark
    & \cmark & \xmark & \xmark
    & \cmark & \xmark & \xmark
    & 1,334 & 3,720 & M & 30.0 \\
    
    TVQA~\citep{lei2018tvqa}
    & \cmark & \xmark & \cmark & \xmark
    & \xmark & \xmark & \cmark
    & \cmark & \xmark & \xmark
    & 2,179 & 15,253 & M & 11.2 \\

    TVQA+~\citep{lei2020tvqa+}
    & \cmark & \xmark & \cmark & \xmark
    & \cmark & \cmark & \cmark
    & \cmark & \xmark & \xmark
    & 403 & 2,821 & M & 61.5 \\

    VidSTG~\citep{zhang2020vidstg}
    & \cmark & \xmark & \cmark & \xmark
    & \cmark & \cmark & \xmark
    & \cmark & \xmark & \xmark
    & 732 & 10,303 & M & 28.0 \\

    NExT-QA~\citep{xiao2021nextqa}
    & \cmark & \xmark & \cmark & \xmark
    & \xmark & \xmark & \cmark
    & \cmark & \xmark & \xmark
    & 1,000 & 8,564 & A & 44.0 \\

    HC-STVG~\citep{tang2021HC-STVG}
    & \cmark & \xmark & \cmark & \xmark
    & \cmark & \cmark & \xmark
    & \cmark & \xmark & \xmark
    & 1,160 & 1,160 & M & 20.0 \\

    QVHighlights~\citep{lei2021QVHighlights}
    & \cmark & \xmark & \cmark & \xmark
    & \cmark & \xmark & \xmark
    & \xmark & \cmark & \xmark
    & 1,529 & 1,542 & M & 149.6 \\

    EgoSchema~\citep{mangalam2023egoschema}
    & \cmark & \xmark & \cmark & \xmark
    & \xmark & \xmark & \cmark
    & \xmark & \cmark & \xmark
    & 5,063 & 5,063 & H & 180.0 \\


    TempCompass~\citep{liu2024tempcompass}
    & \cmark & \xmark & \cmark & \xmark
    & \xmark & \xmark & \cmark
    & \cmark & \xmark & \xmark
    & 410 & 7,540 & H & 11.4 \\

    MVBench~\citep{li2024mvbench}
    & \cmark & \xmark & \cmark & \xmark
    & \xmark & \xmark & \cmark
    & \cmark & \xmark & \xmark
    & 3,641 & 4,000 & A & 16.0 \\

    MLVU~\citep{zhou2024mlvu}
    & \cmark & \xmark & \cmark & \xmark
    & \xmark & \xmark & \cmark
    & \xmark & \cmark & \cmark
    & 1,334 & 2,593 & H & 720.0 \\

    Video-MME~\citep{fu2024videomme}
    & \cmark & \xmark & \cmark & \xmark
    & \xmark & \xmark & \cmark
    & \cmark & \cmark & \cmark
    & 900 & 2,700 & M & 1017.9 \\

    LongVideoBench~\citep{wu2024longvideobench}
    & \cmark & \xmark & \cmark & \xmark
    & \xmark & \xmark & \cmark
    & \cmark & \cmark & \cmark
    & 3,763 & 6,678 & M & 473.0 \\

    CG-Bench~\citep{chen2025cgbench}
    & \cmark & \xmark & \cmark & \xmark
    & \xmark & \xmark & \cmark
    & \xmark & \xmark & \cmark
    & 1,219 & 12,129 & M & 1624.4 \\

    V-STaR~\citep{cheng2025vstar}
    & \cmark & \xmark & \cmark & \xmark
    & \cmark & \cmark & \cmark
    & \cmark & \cmark & \cmark
    & 1,000 & 2,094 & H & 110.2 \\

    IV-Bench~\citep{ma2026ivbench}
    & \xmark & \cmark & \cmark & \xmark
    & \xmark & \xmark & \cmark
    & \xmark & \cmark & \cmark
    & 967 & 2,585 & M & 537.0 \\

    \rowcolor{omniyellow}
    \textbf{OmniCompass-VL (Ours)}
    & \cmark & \cmark & \cmark & \xmark
    & \cmark & \cmark & \cmark
    & \cmark & \cmark & \cmark
    & 820 & 2,460 & M & 354.0 \\

    \midrule
    
    \rowcolor{omnigreen}
    \multicolumn{15}{l}{\textbf{\textit{Audio-Visual Benchmarks}}} \\
    \midrule

    AVUT~\citep{yang2025avut}
    & \cmark & \xmark & \cmark & \cmark
    & \xmark & \xmark & \cmark
    & \cmark & \xmark & \xmark
    & 2,662 & 11,609 & H & 67.8 \\

    Daily-Omni~\citep{zhou2025dailyomni}
    & \cmark & \xmark & \cmark & \cmark
    & \xmark & \xmark & \cmark
    & \cmark & \xmark & \xmark
    & 684 & 1,197 & H & 42.8 \\

    OmniVideoBench~\citep{li2025omnivideobench}
    & \cmark & \xmark & \cmark & \cmark
    & \xmark & \xmark & \cmark
    & \cmark & \cmark & \cmark
    & 628 & 1,000 & H & 384.2 \\

    WorldSense~\citep{hong2025worldsense}
    & \cmark & \xmark & \cmark & \cmark
    & \xmark & \xmark & \cmark
    & \cmark & \cmark & \cmark
    & 1,662 & 3,172 & M & 141.1 \\

    JointAVBench~\citep{chao2026jointavbench}
    & \cmark & \xmark & \cmark & \cmark
    & \xmark & \xmark & \cmark
    & \cmark & \cmark & \cmark
    & 1,046 & 2,853 & H & 97.2 \\

    Video-MME-v2~\citep{fu2026videommev2}
    & \cmark & \xmark & \cmark & \cmark
    & \xmark & \xmark & \cmark
    & \cmark & \cmark & \cmark
    & 800 & 3,200 & M & 624.0 \\
    
    \rowcolor{omniyellow}
    \textbf{OmniCompass-AV (Ours)}
    & \cmark & \cmark & \cmark & \cmark
    & \cmark & \cmark & \cmark
    & \cmark & \cmark & \cmark
    & 240 & 720 & M & 427.5 \\

    \bottomrule
    \end{tabular}
    \vspace{-4mm}
    \end{threeparttable}
    }
\end{table*}

With the rapid development of MLLMs, how to comprehensively evaluate their performance in complex video understanding tasks has become an important problem.
Table~\ref{tab:benchmark_comparison} compares existing benchmarks across four dimensions.
\textbf{(i) Target Specification} defines how targets are presented: through only textual queries~\citep{fu2024videomme} or with reference images~\citep{ma2026ivbench}.
\textbf{(ii) Video Evidence} distinguishes tasks resolved from video frames~\citep{chen2025cgbench} from those requiring synchronized audio-visual content~\citep{hong2025worldsense}.
\textbf{(iii) Task Coverage} specifies the output: temporal grounding (TG) and spatial grounding (SG) predict target-specific intervals and regions~\citep{lei2021QVHighlights,cheng2025vstar}, while question answering (QA) predicts answers~\citep{fu2026videommev2}.
\textbf{(iv) Video Duration} determines the temporal scope: short-video benchmarks emphasize fine-grained motion and local dynamics~\citep{li2024mvbench}, whereas long-video benchmarks test long-range dependencies and information integration across temporally distant events~\citep{zhou2024mlvu}.

Despite this progress, existing benchmarks have three limitations in systematically diagnosing omni-modal video grounding and understanding.
\textbf{(i) Limited video-evidence coverage.}
Benchmarks typically assess either visual information~\citep{chen2025cgbench} or synchronized audio-visual content~\citep{hong2025worldsense}.
In news broadcasts, some queries concern visible scenes; others require aligning spoken reports with people or events.
Both settings are needed to assess visual perception and audio-visual integration.
\textbf{(ii) Inflexible target-specification settings.}
Most benchmarks use only text~\citep{fu2026videommev2}, while IV-Bench combines reference images with text~\citep{ma2026ivbench}.
Practical targets may be specified verbally or with an image, yet these benchmarks do not evaluate the same target under both settings.
\textbf{(iii) Fragmented task coverage.}
Most benchmarks focus on QA~\citep{li2024mvbench} or grounding~\citep{gao2026omniground}; audio-visual benchmarks rarely cover the latter~\citep{li2025omnivideobench,zhou2025dailyomni}.
In basketball games, users may ask \textbf{when} a referenced player scores, \textbf{where} the player is on the court, and \textbf{what} happens.
Joint evaluation distinguishes failures in temporal retrieval, spatial localization, and semantic understanding.

\begin{figure}[t]
    \centering
    \includegraphics[width=\linewidth]{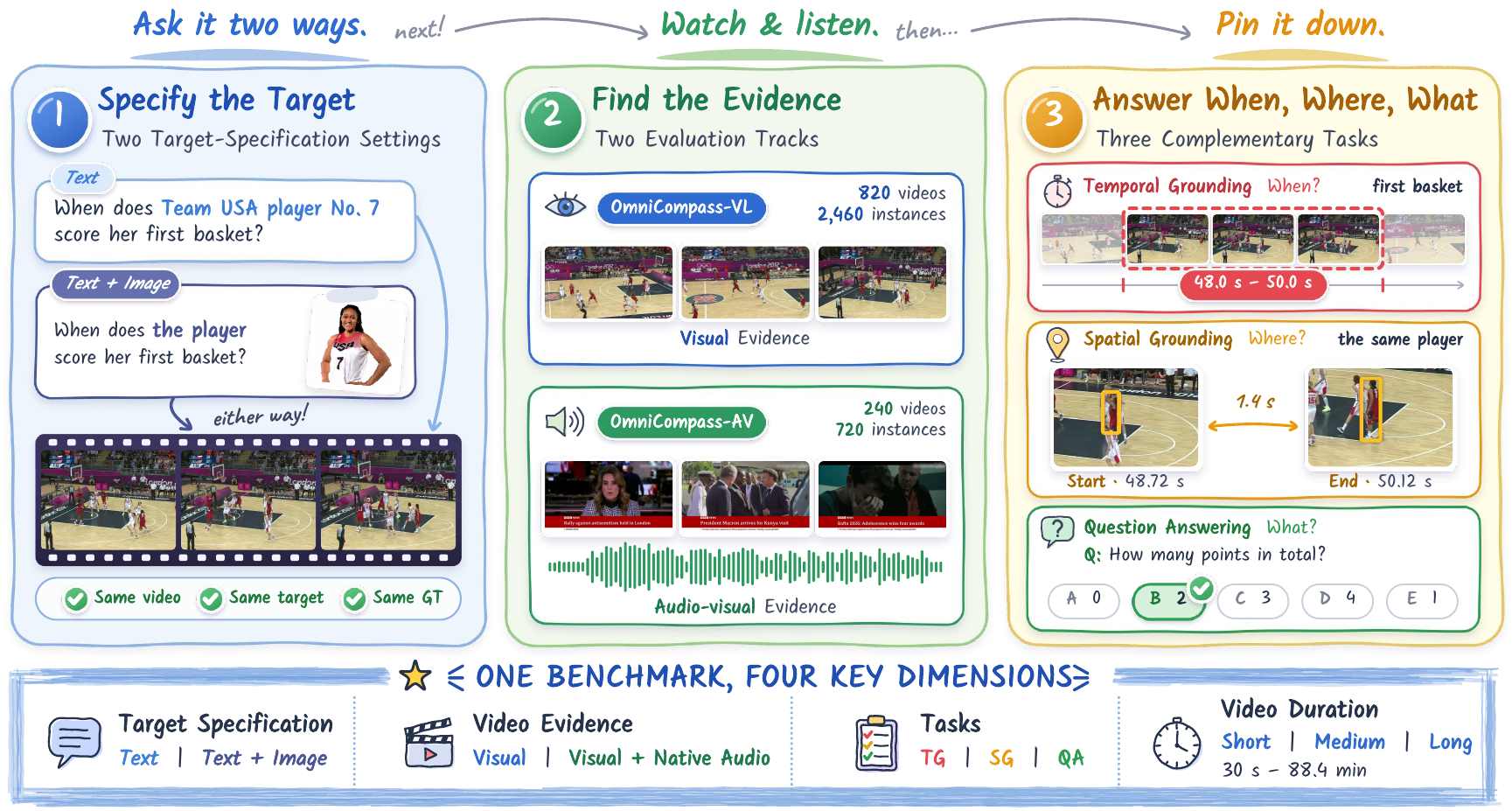}
    \vspace{-5mm}
    \caption{\textbf{Overview of OmniCompass.} The benchmark spans four dimensions:
    target specification, video evidence, task coverage, and video duration.
    Paired \textit{Text}/\textit{Text+Image} specifications are evaluated on the same
    targets. OmniCompass-VL contains 820 videos and 2,460 instances, while the
    240-video OmniCompass-AV subset contains 720 instances with native audio.
    Both tracks cover temporal grounding, spatial grounding, and question answering.}
    \vspace{-5mm}
    \label{fig:omnicompass_overview}
\end{figure}

To address these gaps, we introduce \textbf{OmniCompass}, a holistic benchmark for omni-modal video grounding and understanding.
As shown in Figure~\ref{fig:omnicompass_overview}, OmniCompass has four key features:
\textbf{(i) Complementary Video Evidence.} OmniCompass-VL evaluates understanding from visual content, while OmniCompass-AV additionally requires audio.
\textbf{(ii) Flexible Target Specification.} Each instance pairs \textit{Text} and \textit{Text+Image} settings, specifying the same target through language alone or with a reference image.
The video, task semantics, and ground truth are the same.
We introduce Cross-Specification Consistency Scores (CS) to jointly measure task accuracy and prediction consistency across these settings.
\textbf{(iii) Unified Task Coverage.}
Both tracks provide TG, SG, and QA tasks per video.
The VL track covers five perception and three reasoning categories,
whereas the AV track covers three perception and two reasoning categories.
\textbf{(iv) Diverse Durations and Content.} OmniCompass-VL contains 820 videos and 2,460 instances across short, medium, and long durations.
It combines public sports, traffic, surveillance, and documentary videos with 240 self-collected movie, TV, variety, and news videos.
This subset forms OmniCompass-AV with 720 instances.

We evaluate \textbf{50 MLLMs}, including proprietary and open-source models.
The best proprietary model exceeds the best open-source model by
\textbf{3.2\%} in $\mathrm{CS}^{\mathrm{Avg}}$ on VL track and \textbf{32.8\%} on AV track.
Strong QA performance coexists with weak grounding and cross-reference
inconsistency, even when aggregate accuracies are similar across settings.
Visual target specifications do not consistently help: \textit{Text+Image} improves the single-setting average for only 11 of 36 models.
Further analyses show that timestamps substantially improve model performance across tasks and tracks, while audio consistently improves QA and TG performance on AV track.
Future MLLMs therefore need stronger cross-modal alignment and more reliable audio-aware grounding.
Our contributions are as follows:

\begin{enumerate}[leftmargin=12pt, topsep=2pt, itemsep=2pt]
    \item \textbf{A Holistic Evaluation Framework.}
    OmniCompass unifies target-guided omni-modal video grounding and understanding across four dimensions: paired \textit{Text} and \textit{Text+Image} specifications, VL/AV evidence tracks, TG/SG/QA tasks, and short-to-long videos.
    We introduce CS to jointly assess task correctness and consistency across paired target specifications.

    \item \textbf{A Diverse, Rigorously Verified Benchmark.}
    OmniCompass spans 820 videos, 80.6 hours, seven domains, and 52 subcategories, with 240 videos shared between VL and AV tracks.
    Construction required about 1,800 human-hours for annotator training, calibration, annotation, expert adjudication, and revision.
    All instances undergo multi-round quality verification.

    \item \textbf{A Diagnostic Evaluation of MLLMs.}
    We evaluate 50 different MLLMs, revealing an overall advantage for proprietary models, especially in audio-visual understanding.
    Strong QA performance coexists with weak grounding and cross-specification inconsistency.
    Further analyses examine the effects of target specifications, video duration, timestamps, and input modalities.
\end{enumerate}
\section{Related Work}

\subsection{Multimodal Large Language Models for Video Understanding}

MLLMs connect language models with multimodal encoders for perception and reasoning~\citep{Liu:LLaVA-1.5,Chen:InternVL-v1.5,xu2025qwen2-omni,google2025gemini3,lin2025draw,lin2025PAM,an2026llava-ov2,wang2026locateanything}.
LLaVA-OneVision~\citep{li2024llava-ov} and LLaVA-Video~\citep{zhang2024llava-video} unify image-video modeling.
The latter uses synthetic captions and question-answering data for video instruction tuning.
LLaVA-OneVision-2~\citep{an2026llava-ov2} further adopts OneVision-Encoder~\citep{tang2026ov-encoder} for native-resolution visual encoding and efficient long-video processing through codec-aware token compression.
VideoLLaMA3~\citep{zhang2025videollama} and InternVL3.5~\citep{wang2025internvl3.5} strengthen video training and reasoning.
Qwen3-VL~\citep{Qwen3-VL} improves long-context video understanding through temporal alignment, while Qwen3-Omni~\citep{Qwen3-Omni} jointly processes vision, audio, and language.
Keye-VL-2.0~\citep{team2026keye-vl-2} adapts DSA for efficient long-context processing and consolidates task-specific capabilities through cross-modal MOPD.
TimeLens~\citep{zhang2026timelens} improves temporal grounding with interleaved timestamps and thinking-free reinforcement learning, and TimeLens2~\citep{zhu2026timelens2} further broadens task coverage through verified multi-interval supervision and temporal Wasserstein rewards.
Agentic extensions also explore tool-assisted video reasoning to selectively acquire task-relevant visual and audio evidence~\citep{wang2026videoitg,yang2026longvt,yang2026paravt,meng2025open-o3-video,xing2026omniagent}.
In brief, modern omni-modal video understanding identifies referenced targets and grounds evidence in time and space by integrating visual and audio information.

\subsection{Video Understanding Benchmarks}

Table~\ref{tab:benchmark_comparison} compares existing benchmarks across four design dimensions.
\textbf{(i) Target Specification} defines how the target is presented. Most benchmarks use textual questions or instructions~\citep{fu2024videomme,cheng2025vstar}, while IV-Bench~\citep{ma2026ivbench} pairs text with a reference image for visual specification.
\textbf{(ii) Video Evidence} distinguishes visual and audio-visual requirements. Vision-language benchmarks assess appearance, motion, and temporal relations. TempCompass~\citep{liu2024tempcompass} evaluates temporal perception, while MotionBench~\citep{hong2025motionbench} targets fine-grained motion. WorldSense~\citep{hong2025worldsense} and JointAVBench~\citep{chao2026jointavbench} require synchronized visual and native-audio evidence.
\textbf{(iii) Task Coverage} defines the expected output. Question-answering benchmarks~\citep{fu2024videomme,li2025omnivideobench} produce semantic responses, while grounding benchmarks require explicit evidence localization. OmniGround evaluates temporal intervals and spatial regions~\citep{gao2026omniground}, and V-STaR~\citep{cheng2025vstar} jointly assesses TG, SG, and QA.
\textbf{(iv) Video Duration} sets the temporal scope for evidence integration. MVBench emphasizes local actions and temporal dynamics~\citep{li2024mvbench}, whereas LongVideoBench~\citep{wu2024longvideobench}, MLVU~\citep{zhou2024mlvu}, and LVBench~\citep{wang2025lvbench} stress long-context management across distant moments.
LongVideoBench also evaluates referring reasoning over interleaved video frames and subtitles.
Existing benchmarks cover selected dimensions or fixed combinations, limiting diagnosis of target identification, grounding, and omni-modal reasoning across different target and evidence settings.

OmniCompass unifies these dimensions through paired target-specification settings, complementary evidence tracks, and TG, SG, and QA across video durations. It supports diagnosis of target identification, spatio-temporal grounding, and event understanding under paired reference settings.
\section{The OmniCompass Benchmark}
\label{sec:benchmark}

OmniCompass evaluates omni-modal video grounding and understanding along the dimensions in Table~\ref{tab:benchmark_comparison}.
Figure~\ref{fig:omnicompass_overview} summarizes the evaluation workflow.
Each instance uses paired \textit{Text} and \textit{Text+Image} specifications of the same target.
\textbf{OmniCompass-VL} evaluates vision-language models on visually resolvable tasks, while \textbf{OmniCompass-AV} evaluates omni-modal models using joint visual and native-audio evidence.
Both tracks cover TG, SG, and QA across short (30s $\leq t < 90$s), medium (90s $\leq t < 300$s), and long ($t \geq 300$s) videos.

\subsection{Evaluation Formulation and Metrics}

\noindent\textbf{Two Target-Specification Settings.}
OmniCompass uses paired \textit{Text} and \textit{Text+Image} settings to test whether models consistently localize and understand targets across reference modalities.
Targets include entities, actions, attributes, and scenes.
In \textit{Text}, a textual description $\mathbf{r}^{\text{text}}$ alone specifies the target, such as \textit{``When does the USA player wearing No.~33 complete the dunk?''}.
In \textit{Text+Image}, a textual expression $\mathbf{r}^{\text{text}'}$ and a reference image $\mathbf{r}^{\text{image}}$ jointly specify the target.
For example, \textit{``When does the player complete the dunk?''} requires the image to resolve which player is referenced.
Thus, $\mathbf{r}=\mathbf{r}^{\text{text}}$ or $\mathbf{r}=(\mathbf{r}^{\text{text}'},\mathbf{r}^{\text{image}})$.
Each pair shares the same video, target, task semantics, and ground truth, allowing direct comparison across target specifications.

\noindent\textbf{Two Evaluation Tracks.}
The two tracks share the same target-specification and task contracts but
differ in the video evidence available to the model. Let
$\mathcal{F}_K(\mathcal{V})$ denote $K$ uniformly sampled video frames paired
with their timestamps, and let $\mathcal{A}(\mathcal{V})$ denote the
synchronized audio track. Given a target specification
$\mathbf{r}\in\{\mathbf{r}^{\mathrm{text}},
(\mathbf{r}^{\mathrm{text}'},\mathbf{r}^{\mathrm{image}})\}$, \textbf{OmniCompass-VL} receives
$\mathbf{x}_{\mathrm{VL}}=(\mathcal{F}_K(\mathcal{V}),\mathbf{r})$, whereas
\textbf{OmniCompass-AV} receives
$\mathbf{x}_{\mathrm{AV}}=(\mathcal{F}_K(\mathcal{V}),
\mathcal{A}(\mathcal{V}),\mathbf{r})$. VL queries are answerable from visual
evidence, while AV queries require information from the synchronized audio.

\noindent\textbf{Cross-Specification Consistency Scores (CS).}
Paired \textit{Text} and \textit{Text+Image} settings allow us to assess \textbf{whether a model correctly identifies the same target across referential modalities}.
We introduce \textbf{CS} to jointly measure task correctness and consistency between paired predictions, so agreement alone does not guarantee a high score.
For each task $q\in\{\mathrm{TG},\mathrm{SG},\mathrm{QA}\}$, let $N_q$ denote the number of evaluated pairs, with superscripts $\mathrm{T}$ and $\mathrm{TI}$ denoting the \textit{Text} and \textit{Text+Image} settings.
For TG, let $g_i^{\mathrm{T}}$ and $g_i^{\mathrm{TI}}$ be the tIoUs of the predicted intervals with the ground-truth interval, and $a_i$ the tIoU between the two predictions.
We define $\mathrm{CS}_{\mathrm{TG}}=\frac{1}{N_{\mathrm{TG}}}\sum_{i=1}^{N_{\mathrm{TG}}}\min\{g_i^{\mathrm{T}},g_i^{\mathrm{TI}},a_i\}$.
For SG, let $g_{i,k}^{\mathrm{T}}$ and $g_{i,k}^{\mathrm{TI}}$ be the IoUs between predicted and ground-truth boxes at the start and end boundaries $k\in\{s,e\}$.
Let $a_{i,k}$ denote the IoU between the paired predictions at boundary $k$.
We define $\mathrm{CS}_{\mathrm{SG}}=\frac{1}{2N_{\mathrm{SG}}}\sum_{i=1}^{N_{\mathrm{SG}}}\sum_{k\in\{s,e\}}\min\{g_{i,k}^{\mathrm{T}},g_{i,k}^{\mathrm{TI}},a_{i,k}\}$.
For QA, with predicted answers $\hat{y}_i^{\mathrm{T}}$ and $\hat{y}_i^{\mathrm{TI}}$ and ground-truth answer $y_i$, we define $\mathrm{CS}_{\mathrm{QA}}=\frac{1}{N_{\mathrm{QA}}}\sum_{i=1}^{N_{\mathrm{QA}}}\mathbf{1}\{\hat{y}_i^{\mathrm{T}}=\hat{y}_i^{\mathrm{TI}}=y_i\}$.

\subsection{Benchmark Construction}

\begin{figure}[t]
    \centering
    \includegraphics[width=\linewidth]{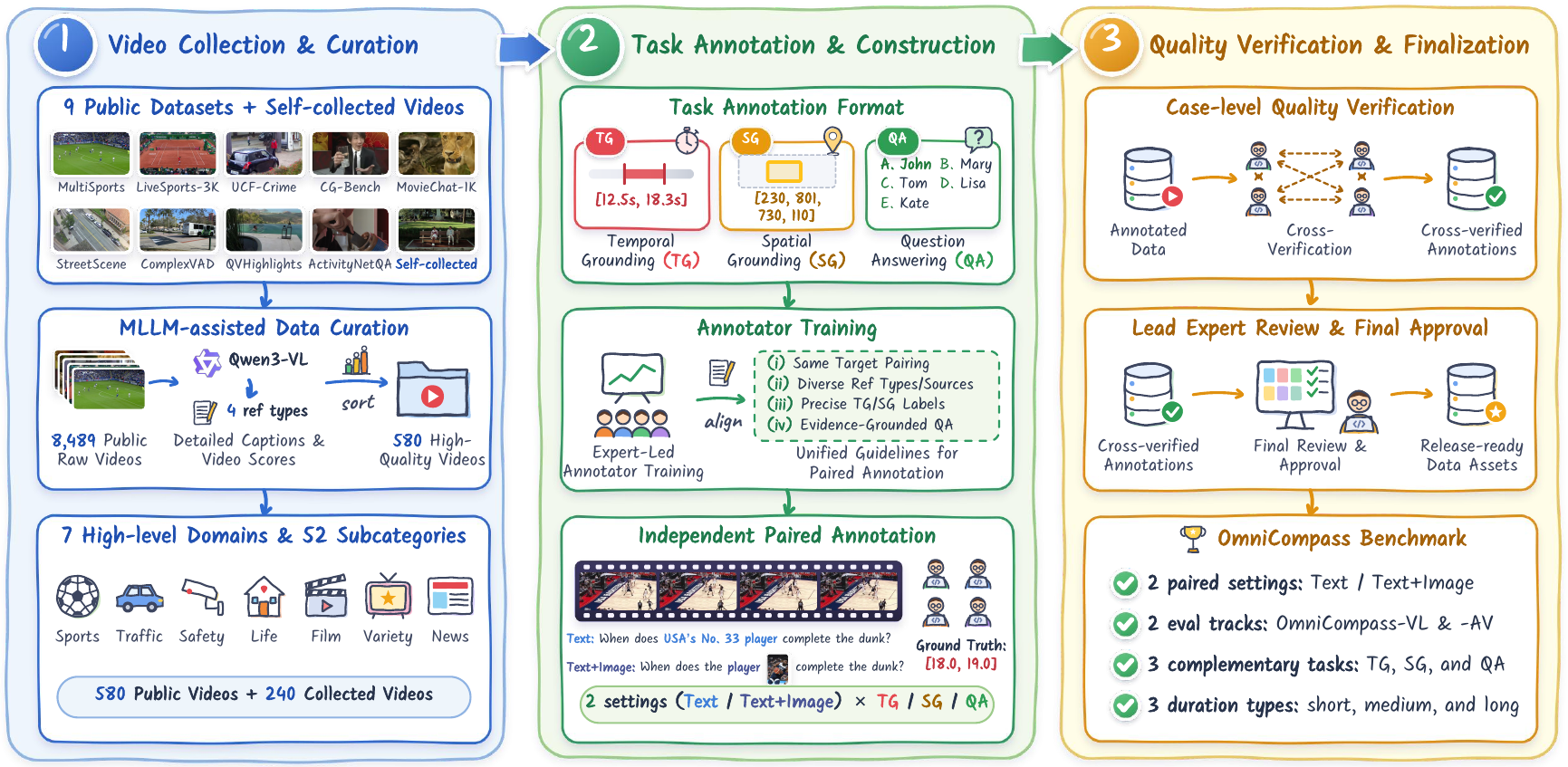}
    \vspace{-5mm}
    \caption{\textbf{Data construction pipeline.} OmniCompass requires about \textbf{1,800 human-hours} for rigorous annotation, verification, and revision.
    \textbf{Stage 1:} We screen 8,489 videos from nine public datasets, retain 6,946 after duration and referential-cue filtering, and select 580 to complement 240 self-collected videos.
    \textbf{Stage 2:} Four trained annotators independently construct paired \textit{Text} and \textit{Text+Image} TG, SG, and QA instances. AV cases include native audio.
    \textbf{Stage 3:} Annotators cross-verify and revise cases before final acceptance and correction by the lead expert, yielding 820 videos across seven domains, paired settings, two tracks, three tasks, and three duration groups.}
    \label{fig:data_construction_pipeline}
    \vspace{-5mm}
\end{figure}

Constructing OmniCompass requires approximately \textbf{1,800 human-hours} for paired annotation, targeted revision, and multi-round quality verification, following the pipeline in Figure~\ref{fig:data_construction_pipeline}.

\noindent\textbf{Stage 1: Video Collection and Curation.}
We construct a pool of \textbf{8,489 candidate videos} from \textbf{nine public datasets} and supplement it with \textbf{240 self-collected videos} from films, TV series, variety shows, and news programs. The resulting pool covers seven high-level domains. Appendix Table~\ref{tab:source_coverage} details the selection scope, characteristics, usage terms, and contributions of the public datasets to the final benchmark.

We use \texttt{Qwen3-VL-8B-Instruct} to profile each public candidate with structured \texttt{entity}, \texttt{action}, \texttt{attribute}, and \texttt{scene} cues. Excluding videos shorter than 30 seconds or missing any of the four referential cue types leaves \textbf{6,946 eligible videos}. We rank them by four predefined criteria: reference necessity, grounding suitability, temporal-spatial utility, and evidence diversity (Appendix Table~\ref{tab:qwen_scoring}). Shortlisting follows domain, duration, source, and subcategory constraints. Manual review retains \textbf{580 high-quality videos}; with 240 self-collected videos, OmniCompass-VL contains 820 videos. The same 240 videos form OmniCompass-AV, with additional audio-visual task annotations. Appendix~\ref{sec:appendix_benchmark_construction} details profiling, filtering, and scoring.

\noindent\textbf{Stage 2: Task Annotation and Construction.}
A lead expert provides two days of task-specific training to four professional annotators before full-scale annotation. The team follows four principles to ensure consistent target specification and evidence across TG, SG, and QA under paired \textit{Text} and \textit{Text+Image} settings.

\textbf{(i) Same-Target Pairing.}
Each \textit{Text} and \textit{Text+Image} pair shares the same video, target, task semantics, and ground truth. In \textit{Text+Image}, the textual expression requires the reference image to identify the target.
\textbf{(ii) Diverse Reference Types and Sources.}
References cover \texttt{entity}, \texttt{action}, \texttt{attribute}, and \texttt{scene} cues using video crops or semantically matched external images. Each image identifies the target without revealing the annotated event or answer evidence. For video-internal references, annotators favor a separate view outside the target interval or QA clue interval whenever available.
\textbf{(iii) Precise TG/SG Labels.}
TG intervals tightly cover complete target-related events, while SG boxes localize the same target at both interval boundaries.
\textbf{(iv) Evidence-Grounded QA Design.}
Following CG-Bench~\citep{chen2025cgbench}, annotators pair a clue interval with five mutually exclusive options: one correct answer and four plausible distractors.

The annotators first conduct a \textbf{pilot annotation} on 30 VL and 30 AV videos spanning all seven domains and three duration types. Each video receives one TG, one SG, and one QA instance under paired \textit{Text} and \textit{Text+Image} settings, yielding \textbf{180 task instances}. The pilot exposes inconsistencies in target pairing, temporal boundaries, spatial target consistency, and QA evidence. The lead expert reviews these cases with the team, resolves ambiguities, and refines the guidelines. The annotators then independently construct the remaining instances under the calibrated protocol. Guideline training, pilot calibration, and initial task construction across both tracks and all three tasks account for approximately \textbf{960 annotator-hours}.

\noindent\textbf{Stage 3: Quality Verification and Finalization.}
All annotated cases undergo case-level quality verification before release. Four annotators independently cross-check annotations from their peers against the Stage 2 principles: same-target pairing, reference diversity and non-leakage, precise TG/SG labels, and evidence-grounded QA design. They inspect temporal-event completeness, boundary-frame box alignment, and the uniqueness and evidential support of QA answers. For AV cases, they also verify native-audio dependence and audio-visual alignment. They correct clear annotation errors, recheck the revised cases, and submit the confirmed cases to the lead expert. The expert resolves remaining inconsistencies, makes necessary corrections, performs final acceptance, and audits the release-ready data. These steps require approximately \textbf{320 annotator-hours} and \textbf{500 lead-expert-hours}. Appendix~\ref{sec:appendix_quality_verification} details the process. Tables~\ref{tab:annotation_quality_control_vl} and~\ref{tab:annotation_quality_control_av} report video-level verification results and typical issues; Table~\ref{tab:annotation_effort} details human effort by stage and role.

\begin{figure}[t]
    \centering
    \includegraphics[width=\linewidth]{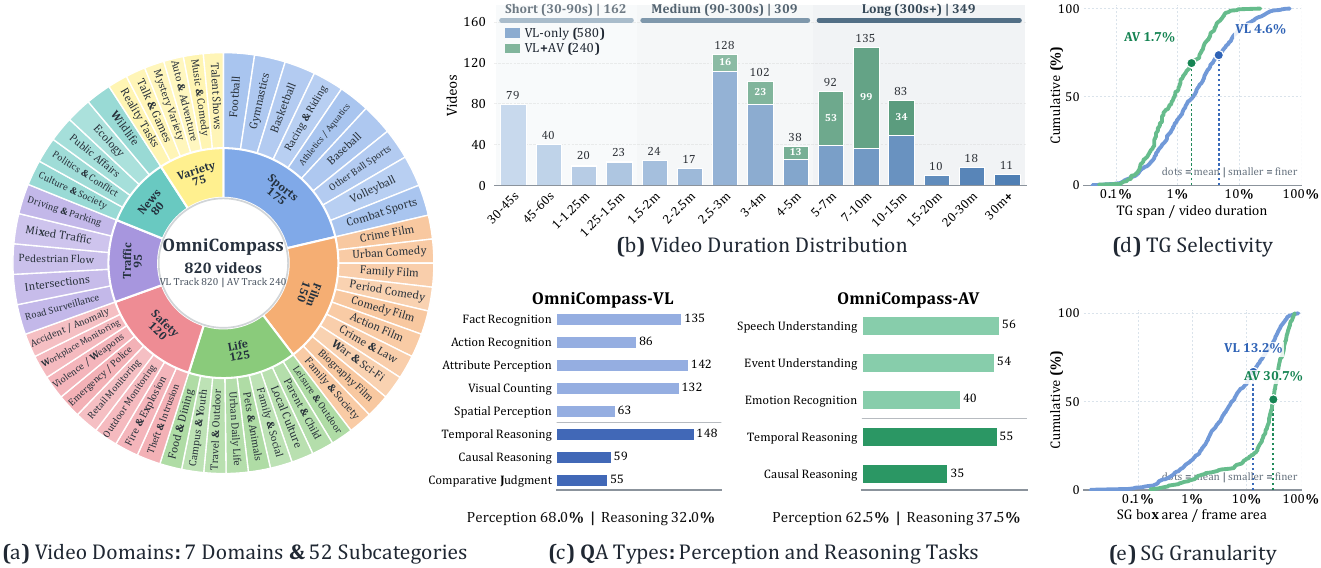}
    \vspace{-5mm}
    \caption{\textbf{Dataset statistics of OmniCompass.}
    \textbf{(a)} 820 videos span 7 domains and 52 subcategories, each with at least ten videos.
    \textbf{(b)} Fifteen duration bins span short (30--90\,s), medium (90--300\,s), and long ($\geq 300$\,s) ranges. Stacked colors distinguish 580 VL-only and 240 shared VL--AV videos.
    \textbf{(c)} Bars report counts for eight OmniCompass-VL and five OmniCompass-AV QA types. Lighter and darker shades distinguish Perception and Reasoning.
    \textbf{(d)} Cumulative TG-span ratios show temporal selectivity.
    \textbf{(e)} Cumulative SG-area ratios show spatial granularity. Dots mark track means.}
    \label{fig:data_statistics}
    \vspace{-5mm}
\end{figure}

\subsection{Benchmark Statistics}

\noindent\textbf{Video Diversity and Temporal Scale.}
OmniCompass contains \textbf{820 videos} across \textbf{7 domains and 52 subcategories}, with 10--43 videos per subcategory and a mean of 15.8.
The videos total \textbf{80.6 hours}, ranging from 30 seconds to 88.4 minutes, with a mean duration of 354.0 seconds.
Figure~\ref{fig:data_statistics}(a,b) shows 162 short, 309 medium, and 349 long videos across 15 duration bins, including 580 VL-only videos and 240 videos shared by both tracks for matched-video evaluation.

\noindent\textbf{Evaluation Breadth and Annotation Granularity.}
Unified TG, SG, and QA annotations yield \textbf{2,460 VL and 720 AV task instances}.
Each video contributes one instance to each task, giving 820 instances per task in VL
and 240 in AV.
Figure~\ref{fig:data_statistics}(c) shows eight VL and five AV QA types.
Perception accounts for 68.0\% and 62.5\%, while reasoning accounts for 32.0\% and
37.5\%, respectively.
The 820 videos span seven domains and 52 subcategories, with 10--43 videos per
subcategory.
QA counts range from 55 to 148 across VL types and from 35 to 56 across AV types,
providing broad coverage without concentrating on a single question category.
Figure~\ref{fig:data_statistics}(d,e) summarizes grounding granularity: TG intervals
average \textbf{4.6\%} of VL and \textbf{1.7\%} of AV video duration, while SG boxes
average \textbf{13.2\%} and \textbf{30.7\%} of frame area, respectively.
Paired \textit{Text} and \textit{Text+Image} settings test target identification,
while the shared AV subset supports matched-video comparisons.
Appendix Figure~\ref{fig:reference_type_source_composition} reports reference types
and image sources across domains.

\section{Experiments}
\label{sec:Experiments}
\subsection{Experimental Setting}
\label{subsec:Setting}

\noindent\textbf{Evaluated Models.}
We evaluate \textbf{50 models} across tracks. 
OmniCompass-VL evaluates vision-language models including proprietary models Gemini
series~\citep{google2025gemini3}, GPT
series~\citep{openai2026gpt55,openai2026gpt56,openai2026gpt6}, Seed
series~\citep{seed2026seed2}, and the Qwen close-source series~\citep{qwen3.8}.
The open-source models cover Kimi-K3~\citep{kimi2026k3},
GLM-5.3-Flash~\citep{zeng2026glm5},
DeepSeek-V4.1-Flash~\citep{deepseekai2026deepseekv41flash}, MiMo-V2.5~\citep{mimov25}, 
the Qwen open-source series~\citep{qwen36_35b_a3b,qwen3.5,Qwen3-VL}, InternVL3.5-38B~\citep{wang2025internvl3.5}, Keye-VL-2.0-30B-A3B~\citep{team2026keye-vl-2}, TimeLens series~\citep{zhang2026timelens,zhu2026timelens2}, and the LLaVA
series~\citep{an2026llava-ov2,zhang2024llava-video}.
OmniCompass-AV evaluates omni-modal
models with native audio support, including the Gemini series, Qwen-Omni series~\citep{team2026qwen3.5omni,qwen3_omni_20251201}, and Seed-2.0-Lite/Mini~\citep{seed2026seed2}, as well as MiMo-V2.5~\citep{mimov25}, video-SALMONN 2+~\citep{tang2025video-SALMONN2}, MiniCPM-o-4.5-8B~\citep{cui2026minicpmo-4.5},
Ola-7B~\citep{liu2025ola},
VideoLLaMA2-7B~\citep{cheng2024videollama2}, and
VITA-1.5-7B~\citep{fu2026vita1.5}. 

\noindent\textbf{Implementation Details.}
\textbf{(i) Frame sampling:}
We uniformly sample 64 frames for proprietary models and open-source models with over 100B total parameters, and 32 for smaller models.
GPT-series models use 49 frames to accommodate the \textit{Text+Image} reference within the 50-image limit.
VITA-1.5-7B follows its official 16-frame limit.
Following TimeLens~\citep{zhang2026timelens}, each frame is immediately preceded by its timestamp in seconds.
\textbf{(ii) Input modalities:}
OmniCompass-VL excludes audio, while OmniCompass-AV adds the native audio following WorldSense~\citep{hong2025worldsense} and Video-MME-v2~\citep{fu2026videommev2}.
In \textit{Text+Image}, the reference image follows these inputs and immediately precedes the task prompt.
Appendix~\ref{sec:appendix_eval_details} details the experimental setup, input serialization, and task-specific prompts and response formats.

\begin{table*}[!t]
\caption{
\textbf{Results on OmniCompass-VL.}
We evaluate TG, SG, and QA without audio using uniformly sampled,
timestamped frames.
Performance is reported as \textit{Text}/\textit{Text+Image}, while \textbf{CS} requires
paired predictions to be correct and consistent.
\textbf{Average} is the mean of TG m\textsubscript{tIoU}, SG
m\textsubscript{sIoU}, and QA Accuracy per specification, while
CS\textsuperscript{Avg} averages the three CS scores.
Within each group, the best and second-best results are
\textbf{bold} and \underline{underlined}, respectively.
}
\vspace{-2mm}
\label{tab:results_vl}
\centering
\newcommand{\tablemodellogo}[1]{%
  \raisebox{0.35em}[0pt][0pt]{\makebox[1.15em][c]{\raisebox{-0.5\height}{#1}}}\hspace{0.2em}%
}
\setlength{\tabcolsep}{4pt}
\resizebox{\textwidth}{!}{%
\begin{tabular}{
lc|
cc>{\columncolor[rgb]{.949,.949,.949}}c|
cc>{\columncolor[rgb]{.949,.949,.949}}c|
c>{\columncolor[rgb]{.949,.949,.949}}c|
c>{\columncolor[rgb]{.949,.949,.949}}c
}
\toprule
\multirow{2}{*}{\textbf{Models}} &
\multirow{2}{*}{\textbf{Frames}} &
\multicolumn{3}{c|}{\textbf{Temporal Grounding}} &
\multicolumn{3}{c|}{\textbf{Spatial Grounding}} &
\multicolumn{2}{c|}{\textbf{Question Answering}} &
\multicolumn{2}{c}{\textbf{Overall}} \\
\cmidrule(lr){3-5}
\cmidrule(lr){6-8}
\cmidrule(lr){9-10}
\cmidrule(lr){11-12}
& &
\textbf{R.@0.5} &
\textbf{m\textsubscript{tIoU}} &
\textbf{CS\textsuperscript{TG}} &
\textbf{AP@0.5} &
\textbf{m\textsubscript{sIoU}} &
\textbf{CS\textsuperscript{SG}} &
\textbf{Accuracy} &
\textbf{CS\textsuperscript{QA}} &
\textbf{Average} &
\textbf{CS\textsuperscript{Avg}} \\
\midrule

\rowcolor{omniyellow}
\multicolumn{12}{l}{\textbf{\textit{Proprietary MLLMs}}} \\
\midrule

\tablemodellogo{\includegraphics[width=1.1em,height=0.95em,keepaspectratio]{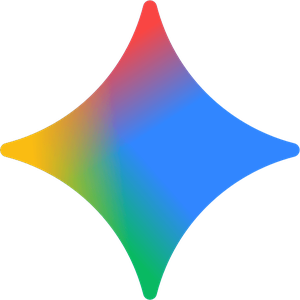}}Gemini-3.8-Flash & 64 &
\textbf{33.8}/\textbf{32.7} &
\textbf{32.6}/\textbf{33.5} &
\textbf{27.1} &
\underline{37.5}/\textbf{41.6} &
37.7/\textbf{41.8} &
32.3 &
73.3/\textbf{76.6} &
\textbf{69.1} &
\textbf{47.9}/\textbf{50.6} &
\textbf{42.8} \\

\tablemodellogo{\includegraphics[width=1.1em,height=0.95em,keepaspectratio]{Figures/model_logos/gemini.png}}Gemini-3.1-Pro & 64 &
24.9/24.9 & 25.6/26.6 & 19.3 &
34.9/35.2 & 36.5/37.3 & 31.8 &
71.4/73.6 & 65.7 & 44.5/45.9 & 38.9 \\

\tablemodellogo{\includegraphics[width=1.1em,height=0.95em,keepaspectratio]{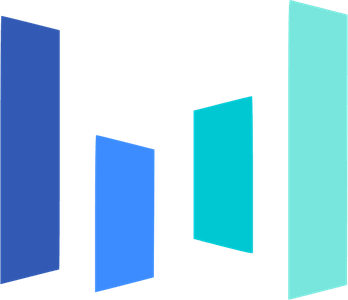}}Seed-2.1-Pro-260915 & 64 &
27.9/29.8 & 29.3/29.9 & 23.7 &
\textbf{40.9}/\underline{38.9} &
\textbf{40.6}/\underline{39.6} &
\textbf{34.9} &
73.4/\underline{74.3} &
67.4 &
\underline{47.8}/\underline{47.9} &
\underline{42.0} \\

\tablemodellogo{\includegraphics[width=1.1em,height=0.95em,keepaspectratio]{Figures/model_logos/bytedance.png}}Seed-2.1-Turbo-260628 & 64 &
29.9/29.4 & 30.5/30.2 & 25.0 &
35.4/35.4 & 36.3/36.3 & 30.0 &
\underline{74.1}/74.1 &
\underline{67.7} &
47.0/46.9 & 40.9 \\

\tablemodellogo{\includegraphics[width=1.1em,height=0.95em,keepaspectratio]{Figures/model_logos/bytedance.png}}Seed-2.0-Pro-260215 & 64 &
\underline{30.7}/\underline{31.0} &
30.6/30.9 & 24.9 &
30.6/31.7 & 34.9/35.3 & 29.4 &
\textbf{74.5}/73.0 &
67.3 & 46.6/46.4 & 40.5 \\

\tablemodellogo{\includegraphics[width=1.1em,height=0.95em,keepaspectratio]{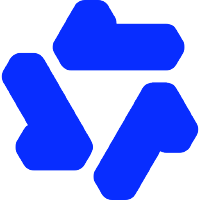}}Qwen3.8-Max & 64 &
29.3/30.3 &
30.7/\underline{32.2} &
\underline{26.7} &
30.1/32.1 & 33.9/34.0 & 26.7 &
72.4/72.5 & 67.2 & 45.7/46.2 & 40.2 \\

\tablemodellogo{\includegraphics[width=1.1em,height=0.95em,keepaspectratio]{Figures/model_logos/qwen.png}}Qwen3.8-Flash & 64 &
28.5/30.1 & 29.5/30.6 & 24.1 &
22.6/24.4 & 28.8/29.3 & 22.6 &
69.2/71.0 & 62.4 & 42.5/43.6 & 36.4 \\

\tablemodellogo{\includegraphics[width=1.1em,height=0.95em,keepaspectratio]{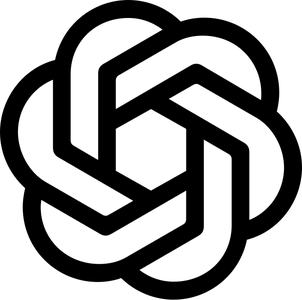}}GPT-6-Astra & 49 &
30.0/30.3 &
\underline{30.9}/31.7 &
26.3 &
34.6/35.9 &
\underline{37.8}/38.3 &
\underline{32.7} &
69.4/65.6 & 59.4 & 46.0/45.2 & 39.5 \\

\tablemodellogo{\includegraphics[width=1.1em,height=0.95em,keepaspectratio]{Figures/model_logos/openai.png}}GPT-5.6-Sol & 49 &
25.2/28.7 & 27.2/30.5 & 22.2 &
31.7/29.8 & 34.8/33.8 & 28.0 &
69.5/69.1 & 62.4 & 43.8/44.5 & 37.5 \\

\tablemodellogo{\includegraphics[width=1.1em,height=0.95em,keepaspectratio]{Figures/model_logos/openai.png}}GPT-5.6-Terra & 49 &
20.1/23.7 & 24.1/26.3 & 19.3 &
19.4/20.9 & 26.3/26.6 & 20.1 &
68.4/69.5 & 62.0 & 39.6/40.8 & 33.8 \\

\tablemodellogo{\includegraphics[width=1.1em,height=0.95em,keepaspectratio]{Figures/model_logos/openai.png}}GPT-5.6-Luna & 49 &
21.0/21.3 & 24.7/24.8 & 19.2 &
27.3/28.9 & 32.2/33.1 & 26.8 &
65.5/64.6 & 56.7 & 40.8/40.8 & 34.2 \\

\tablemodellogo{\includegraphics[width=1.1em,height=0.95em,keepaspectratio]{Figures/model_logos/openai.png}}GPT-5.5 & 49 &
23.3/24.5 & 26.0/27.5 & 21.3 &
20.6/20.5 & 26.4/26.3 & 20.7 &
67.9/70.2 & 62.7 & 40.1/41.3 & 34.9 \\
\midrule
\rowcolor{omniblue}
\multicolumn{12}{l}{\textbf{\textit{Open-Source MLLMs}}} \\
\midrule

\tablemodellogo{\includegraphics[width=1.1em,height=0.95em,keepaspectratio]{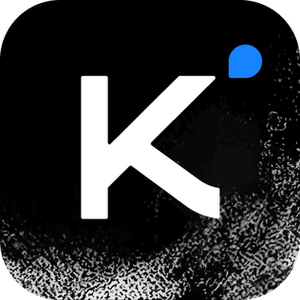}}Kimi-K3 & 64 & 23.5/24.6 & 27.1/27.7 & 22.6 &
\textbf{30.0}/\textbf{31.6} &
\underline{33.3}/\textbf{35.2} &
\textbf{28.4} &
\textbf{73.0}/\textbf{72.9} &
\textbf{67.6} &
\textbf{44.5}/\textbf{45.3} &
\textbf{39.6} \\

\tablemodellogo{\includegraphics[width=1.1em,height=0.95em,keepaspectratio]{Figures/model_logos/qwen.png}}Qwen3.5-397B-A17B & 64 &
\textbf{28.9}/\textbf{28.5} &
\underline{28.6}/\underline{28.7} &
\underline{23.1} &
\underline{28.2}/\underline{29.6} &
\textbf{33.4}/\underline{33.7} &
\underline{28.0} &
67.7/\underline{69.6} &
60.6 &
\underline{43.3}/\underline{44.0} &
\underline{37.2} \\

\tablemodellogo{\includegraphics[width=1.1em,height=0.95em,keepaspectratio]{Figures/model_logos/qwen.png}}Qwen3.5-122B-A10B & 64 &
\underline{27.2}/\underline{27.2} &
\textbf{29.7}/\textbf{29.1} &
\textbf{24.3} &
20.0/19.4 & 25.2/25.2 & 20.8 & 62.6/62.5 & 55.4 &
39.2/38.9 & 33.5 \\

\tablemodellogo{\includegraphics[width=1.1em,height=0.95em,keepaspectratio]{Figures/model_logos/qwen.png}}Qwen3-VL-235B-A22B-Instruct & 64 &
27.0/25.4 & 28.5/26.0 & 21.6 &
18.6/18.8 & 26.0/25.0 & 20.9 &
\underline{68.4}/65.3 &
\underline{61.0} &
41.0/38.8 & 34.5 \\

\tablemodellogo{\includegraphics[width=1.1em,height=0.95em,keepaspectratio]{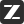}}GLM-5.3-Flash & 64 & 24.4/23.0 & 25.5/24.4 & 18.3 &
15.7/14.6 & 22.9/23.5 & 14.6 & 68.3/68.7 & 59.4 &
38.9/38.9 & 30.8 \\

\tablemodellogo{\includegraphics[width=1.1em,height=0.95em,keepaspectratio]{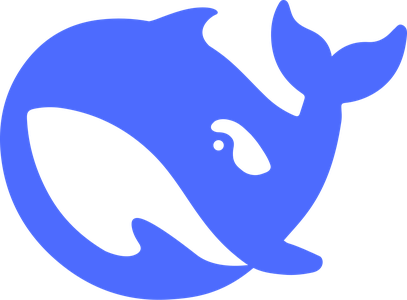}}DeepSeek-V4.1-Flash & 64 & 15.7/17.4 & 19.1/19.4 & 12.4 &
11.5/11.6 & 18.1/17.9 & 10.5 & 66.2/64.6 & 56.6 &
34.5/34.0 & 26.5 \\

\tablemodellogo{\includegraphics[width=1.1em,height=0.95em,keepaspectratio]{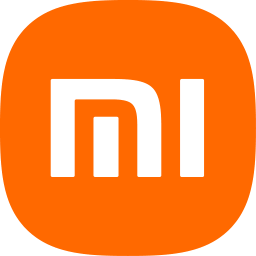}}MiMo-V2.5 & 64 & 17.8/18.6 & 21.6/20.8 & 13.8 &
14.2/15.3 & 21.9/22.3 & 13.6 & 62.0/60.8 & 49.0 &
35.2/34.7 & 25.5 \\

\tablemodellogo{\includegraphics[width=1.1em,height=0.95em,keepaspectratio]{Figures/model_logos/qwen.png}}Qwen3.8-27B & 32 & 17.5/19.1 & 20.3/22.2 & 16.0 &
16.2/19.0 & 23.9/25.4 & 17.2 & 64.8/62.6 & 55.7 &
36.3/36.7 & 29.6 \\

\tablemodellogo{\includegraphics[width=1.1em,height=0.95em,keepaspectratio]{Figures/model_logos/qwen.png}}Qwen3.6-35B-A3B & 32 & 18.7/18.8 & 19.8/19.7 & 14.4 &
20.9/19.1 & 26.8/26.4 & 20.9 & 59.5/59.9 & 50.2 &
35.4/35.3 & 28.5 \\

\tablemodellogo{\includegraphics[width=1.1em,height=0.95em,keepaspectratio]{Figures/model_logos/qwen.png}}Qwen3.6-27B & 32 & 20.5/18.8 & 21.5/20.4 & 15.7 &
14.4/15.8 & 20.7/22.2 & 15.3 & 62.5/61.2 & 54.8 &
34.9/34.6 & 28.6 \\

\tablemodellogo{\includegraphics[width=1.1em,height=0.95em,keepaspectratio]{Figures/model_logos/qwen.png}}Qwen3.5-35B-A3B & 32 & 17.3/16.4 & 19.5/18.7 & 13.7 &
13.3/13.4 & 21.2/21.9 & 15.8 & 61.3/60.9 & 51.9 &
34.0/33.8 & 27.1 \\

\tablemodellogo{\includegraphics[width=1.1em,height=0.95em,keepaspectratio]{Figures/model_logos/qwen.png}}Qwen3.5-27B & 32 & 19.4/18.0 & 22.0/21.1 & 17.4 &
11.3/11.5 & 18.6/19.1 & 15.3 & 61.5/58.4 & 53.9 &
34.0/32.9 & 28.9 \\

\tablemodellogo{\includegraphics[width=1.1em,height=0.95em,keepaspectratio]{Figures/model_logos/qwen.png}}Qwen3.5-9B & 32 & 17.4/14.4 & 20.2/16.6 & 13.3 &
16.8/16.8 & 22.3/22.3 & 17.7 & 58.5/55.2 & 50.1 &
33.7/31.4 & 27.0 \\

\tablemodellogo{\includegraphics[width=1.1em,height=0.95em,keepaspectratio]{Figures/model_logos/qwen.png}}Qwen3.5-4B & 32 & 16.6/12.4 & 19.5/15.3 & 12.6 &
13.0/11.6 & 18.9/17.1 & 13.8 & 55.7/54.0 & 46.6 &
31.4/28.8 & 24.3 \\

\tablemodellogo{\includegraphics[width=1.1em,height=0.95em,keepaspectratio]{Figures/model_logos/qwen.png}}Qwen3-VL-30B-A3B-Instruct & 32 &
16.8/12.4 & 19.5/14.9 & 11.7 &
15.2/12.9 & 22.1/19.6 & 15.6 & 54.1/49.6 & 43.4 &
31.9/28.0 & 23.6 \\

\tablemodellogo{\includegraphics[width=1.1em,height=0.95em,keepaspectratio]{Figures/model_logos/qwen.png}}Qwen3-VL-8B-Instruct & 32 &
13.2/11.1 & 16.3/13.9 & 11.3 &
13.8/12.4 & 20.4/19.3 & 15.9 & 59.9/56.7 & 51.2 &
32.2/29.9 & 26.1 \\

\tablemodellogo{\includegraphics[width=1.1em,height=0.95em,keepaspectratio]{Figures/model_logos/qwen.png}}Qwen3-VL-4B-Instruct & 32 &
13.8/10.5 & 16.8/13.0 & 10.3 &
6.6/5.6 & 13.4/12.2 & 9.3 & 57.3/56.0 & 50.2 &
29.2/27.1 & 23.3 \\

\tablemodellogo{\includegraphics[width=1.1em,height=0.95em,keepaspectratio]{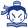}}InternVL3.5-38B & 32 &
13.0/10.5 & 16.3/13.6 & 10.1 &
0.2/0.1 & 2.5/2.5 & 1.4 & 61.3/58.4 & 52.4 &
26.7/24.8 & 21.3 \\

\tablemodellogo{\includegraphics[width=1.1em,height=0.95em,keepaspectratio]{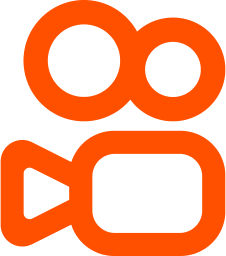}}Keye-VL-2.0-30B-A3B & 32 &
13.4/11.5 & 15.8/13.7 & 11.2 &
2.7/3.3 & 11.5/12.2 & 8.2 & 60.2/61.8 & 53.0 &
29.2/29.2 & 24.1 \\

\tablemodellogo{\includegraphics[width=1.1em,height=0.95em,keepaspectratio]{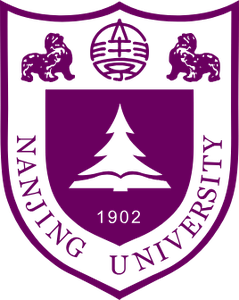}}TimeLens2-8B & 32 &
20.7/18.2 & 20.8/18.5 & 15.3 &
2.4/2.6 & 9.1/8.0 & 6.3 & 59.5/57.0 & 51.0 &
29.8/27.8 & 24.2 \\

\tablemodellogo{\includegraphics[width=1.1em,height=0.95em,keepaspectratio]{Figures/model_logos/timelens.png}}TimeLens-8B & 32 &
22.4/18.5 & 21.4/18.1 & 15.5 &
2.4/2.8 & 8.2/8.7 & 5.9 & 57.7/55.5 & 49.5 &
29.1/27.4 & 23.6 \\

\tablemodellogo{\includegraphics[width=1.1em,height=0.95em,keepaspectratio]{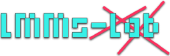}}LLaVA-OneVision-2-8B & 32 &
15.4/10.7 & 15.8/11.2 & 8.4 &
0.9/1.1 & 7.1/6.7 & 4.6 & 57.9/52.4 & 48.0 &
27.0/23.4 & 20.4 \\

\tablemodellogo{\includegraphics[width=1.1em,height=0.95em,keepaspectratio]{Figures/model_logos/llava.png}}LLaVA-Video-7B & 32 &
1.6/2.1 & 4.3/3.1 & 1.5 &
0.0/0.0 & 0.5/0.5 & 0.1 & 55.0/50.7 & 45.6 &
19.9/18.1 & 15.7 \\

\tablemodellogo{\includegraphics[width=1.1em,height=0.95em,keepaspectratio]{Figures/model_logos/llava.png}}LLaVA-OneVision-7B & 32 &
2.6/1.7 & 4.6/2.4 & 1.4 &
0.0/0.0 & 2.0/1.6 & 0.8 & 50.7/47.9 & 42.3 &
19.1/17.3 & 14.8 \\
\bottomrule
\end{tabular}%
}
\vspace{-3mm}
\end{table*}

\subsection{Main Results}
\label{subsec:comparisons}

Tables~\ref{tab:results_vl} and~\ref{tab:results_av} report TG, SG, and QA evaluations under \textit{Text} and \textit{Text+Image}.
Figure~\ref{fig:qa_performance} provides category-wise QA comparisons across selected models in both tracks.
Appendix~\ref{sec:appendix_additional_comparisons} provides additional results across video duration, domain, and target granularity.

\noindent

\noindent\textbf{OmniCompass-VL.}
Table~\ref{tab:results_vl} presents that Gemini-3.8-Flash achieves the highest $\mathrm{CS}^{\mathrm{Avg}}$ score of \textbf{42.8\%}, while Kimi-K3 leads the open-source models with a score of \textbf{39.6\%}.
Experimental results reveal the following insights.
\textbf{(i) Grounding and Fine-Grained Perception Lag Behind Reasoning.}
29 of 36 models exceed 50\% in $\mathrm{CS}^{\mathrm{QA}}$, peaking at 69.1\%. The best $\mathrm{CS}^{\mathrm{TG}}$ and $\mathrm{CS}^{\mathrm{SG}}$ are only 27.1\% and 34.9\%.
Figure~\ref{fig:qa_performance}(a) shows that Visual Counting ranks last in $\mathrm{CS}^{\mathrm{QA}}$ for all eight plotted models, at 26.4\%--48.1\%.
Spatial Perception averages 61.0\%, below 69.8\% for Causal Reasoning.
\textbf{(ii) Visual References Benefit Proprietary Models and Grounding.}
\textit{Text+Image} improves average performance for 11 of the 36 models: 8 of 12 proprietary models and 3 of 24 open-source models.
Among the open-source models with fewer than 100B parameters, only Qwen3.8-27B improves.
It gains 1.9\% and 1.5\% in TG m\textsubscript{tIoU} and SG m\textsubscript{sIoU} but loses 2.2\% in QA accuracy.
Visual cues may aid target matching while omitting useful textual detail.
\textbf{(iii) Larger Models Do Not Lead Every Task.}
Within Qwen3.5, 397B exceeds 122B in $\mathrm{CS}^{\mathrm{Avg}}$, 37.2\% versus 33.5\%, but trails in $\mathrm{CS}^{\mathrm{TG}}$, 23.1\% versus 24.3\%. Likewise, 27B trails 9B in $\mathrm{CS}^{\mathrm{SG}}$, 15.3\% versus 17.7\%.
\textbf{(iv) Cross-Reference Consistency Remains Limited.}
Across 36 models, $\mathrm{CS}^{\mathrm{TG}}$, $\mathrm{CS}^{\mathrm{SG}}$, and $\mathrm{CS}^{\mathrm{QA}}$ average 4.3\%, 4.6\%, and 6.4\% below the lower corresponding single-setting score.
Strong single-setting results do not ensure correctness across settings.

\par
\clearpage\begin{table*}[!t]
\caption{
\textbf{Results on the OmniCompass-AV.}
Models are evaluated on TG, SG, and QA using uniformly sampled, timestamped frames with synchronized audio.
}
\vspace{-2mm}
\label{tab:results_av}
\centering
\newcommand{\tablemodellogo}[1]{%
  \raisebox{0.35em}[0pt][0pt]{\makebox[1.15em][c]{\raisebox{-0.5\height}{#1}}}\hspace{0.2em}%
}
\setlength{\tabcolsep}{4pt}
\resizebox{\textwidth}{!}{%
\begin{tabular}{
lc|
cc>{\columncolor[rgb]{.949,.949,.949}}c|
cc>{\columncolor[rgb]{.949,.949,.949}}c|
c>{\columncolor[rgb]{.949,.949,.949}}c|
c>{\columncolor[rgb]{.949,.949,.949}}c
}
\toprule
\multirow{2}{*}{\textbf{Models}} &
\multirow{2}{*}{\textbf{Frames}} &
\multicolumn{3}{c|}{\textbf{Temporal Grounding}} &
\multicolumn{3}{c|}{\textbf{Spatial Grounding}} &
\multicolumn{2}{c|}{\textbf{Question Answering}} &
\multicolumn{2}{c}{\textbf{Overall}} \\
\cmidrule(lr){3-5}
\cmidrule(lr){6-8}
\cmidrule(lr){9-10}
\cmidrule(lr){11-12}
& &
\textbf{R.@0.5} &
\textbf{m\textsubscript{tIoU}} &
\textbf{CS\textsuperscript{TG}} &
\textbf{AP@0.5} &
\textbf{m\textsubscript{sIoU}} &
\textbf{CS\textsuperscript{SG}} &
\textbf{Accuracy} &
\textbf{CS\textsuperscript{QA}} &
\textbf{Average} &
\textbf{CS\textsuperscript{Avg}} \\
\midrule

\rowcolor{omniyellow}
\multicolumn{12}{l}{\textbf{\textit{Proprietary Omni-modal MLLMs}}} \\
\midrule

\tablemodellogo{\includegraphics[width=1.1em,height=0.95em,keepaspectratio]{Figures/model_logos/gemini.png}}Gemini-3.8-Flash & 64 & \textbf{46.9/48.1} & \textbf{46.4/45.9} & \textbf{40.6} & \textbf{62.8/63.2} & \textbf{59.8/59.3} & \textbf{53.5} & \textbf{80.3/80.8} & \textbf{78.2} & \textbf{62.2/62.0} & \textbf{57.4} \\
\tablemodellogo{\includegraphics[width=1.1em,height=0.95em,keepaspectratio]{Figures/model_logos/gemini.png}}Gemini-3.1-Pro & 64 & 15.1/17.6 & 16.6/17.7 & 9.0 & 54.8/56.1 & 52.0/52.1 & 45.8 & 72.4/71.1 & 64.9 & 47.0/47.0 & 39.9 \\

\tablemodellogo{\includegraphics[width=1.1em,height=0.95em,keepaspectratio]{Figures/model_logos/qwen.png}}Qwen3.8-Omni-Flash & 64 & \underline{26.3/28.5} & \underline{29.2/30.9} & \underline{21.8} & 60.1/59.5 & \underline{56.8/56.0} & \underline{50.9} & \underline{75.9/78.9} & \underline{73.2} & \underline{54.0/55.3} & \underline{48.6} \\

\tablemodellogo{\includegraphics[width=1.1em,height=0.95em,keepaspectratio]{Figures/model_logos/qwen.png}}Qwen3.5-Omni-Plus & 64 & 18.0/14.9 & 21.0/21.1 & 14.0 & \underline{61.8/60.8} & 55.5/55.6 & 50.5 & 71.1/73.2 & 68.4 & 49.2/50.0 & 44.3 \\

\tablemodellogo{\includegraphics[width=1.1em,height=0.95em,keepaspectratio]{Figures/model_logos/qwen.png}}Qwen3.5-Omni-Flash & 64 & 6.7/7.9 & 10.6/10.1 & 5.7 & 49.2/51.3 & 44.2/47.0 & 34.7 & 63.6/64.0 & 58.0 & 39.5/40.4 & 32.8 \\

\tablemodellogo{\includegraphics[width=1.1em,height=0.95em,keepaspectratio]{Figures/model_logos/bytedance.png}}Seed-2.0-Lite-260428 & 64 & 13.3/8.3 & 16.9/13.2 & 10.6 & 56.3/52.9 & 53.3/51.8 & 45.5 & 71.3/69.2 & 63.8 & 47.1/44.7 & 39.9 \\

\tablemodellogo{\includegraphics[width=1.1em,height=0.95em,keepaspectratio]{Figures/model_logos/bytedance.png}}Seed-2.0-Mini-260428 & 64 & 4.2/3.3 & 6.4/3.8 & 2.1 & 27.1/25.4 & 28.6/28.8 & 16.5 & 63.8/60.0 & 53.3 & 32.9/30.8 & 24.0 \\

\midrule
\rowcolor{omnigreen}
\multicolumn{12}{l}{\textbf{\textit{Open-Source Omni-modal MLLMs}}} \\
\midrule

\tablemodellogo{\includegraphics[width=1.1em,height=0.95em,keepaspectratio]{Figures/model_logos/xiaomi.png}}MiMo-V2.5 & 64 & \textbf{6.3}/\textbf{8.6} & \textbf{9.4}/\textbf{8.7} & \underline{3.1} & \textbf{29.2}/\textbf{33.2} & \textbf{34.9}/\textbf{36.0} & \textbf{23.4} & \textbf{59.2}/\textbf{58.7} & 47.3 & \textbf{34.5}/\textbf{34.5} & \textbf{24.6} \\

\tablemodellogo{\includegraphics[width=1.1em,height=0.95em,keepaspectratio]{Figures/model_logos/qwen.png}}Qwen3-Omni-30B-A3B & 32 & \underline{3.8}/\underline{3.3} & \underline{5.5}/\underline{5.0} & \textbf{3.5} & \underline{24.2}/\underline{24.6} & \underline{27.4}/\underline{29.8} & \underline{17.3} & 53.3/\underline{55.0} & \underline{47.9} & \underline{28.7}/\underline{30.0} & \underline{22.9} \\

\tablemodellogo{\includegraphics[width=1.1em,height=0.95em,keepaspectratio]{Figures/model_logos/qwen.png}}Qwen2.5-Omni-7B & 32 & 1.7/0.0 & 1.3/0.6 & 0.2 & 2.1/12.1 & 11.7/22.4 & 7.0 & 47.5/50.0 & 43.8 & 20.1/24.3 & 17.0 \\

\tablemodellogo{\includegraphics[width=1.1em,height=0.95em,keepaspectratio]{Figures/model_logos/qwen.png}}Qwen2.5-Omni-3B & 32 & 1.3/2.1 & 1.4/2.5 & 0.8 & 0.4/2.5 & 1.9/7.6 & 0.7 & 48.8/50.0 & 45.0 & 17.4/20.0 & 15.5 \\

\tablemodellogo{\includegraphics[width=1.1em,height=0.95em,keepaspectratio]{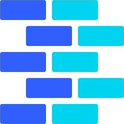}}MiniCPM-o-4.5-8B & 32 & 0.0/0.0 & 1.9/1.4 & 0.4 & 1.3/1.7 & 10.4/9.0 & 5.0 & 52.1/52.9 & 47.1 & 21.5/21.1 & 17.5 \\

\tablemodellogo{\includegraphics[width=1.1em,height=0.95em,keepaspectratio]{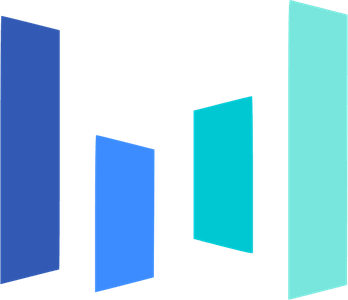}}video-SALMONN 2+ 72B & 32 & \underline{3.8}/1.3 & 5.4/2.3 & 1.6 & 0.0/0.0 & 1.9/2.3 & 1.0 & \underline{56.3}/54.2 & \textbf{50.8} & 21.2/19.6 & 17.8 \\

\tablemodellogo{\includegraphics[width=1.1em,height=0.95em,keepaspectratio]{Figures/model_logos/salmonn.png}}video-SALMONN 2+ 7B & 32 & 2.9/2.1 & 3.1/1.8 & 1.0 & 0.0/0.0 & 0.9/0.1 & 0.1 & 47.1/51.3 & 42.9 & 17.0/17.7 & 14.7 \\

\tablemodellogo{\includegraphics[width=1.1em,height=0.95em,keepaspectratio]{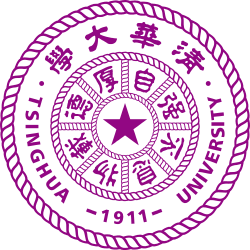}}Ola-7B & 32 & 1.7/0.8 & 2.2/1.3 & 0.0 & 1.7/3.3 & 6.4/10.3 & 2.7 & 42.1/42.5 & 37.5 & 16.9/18.0 & 13.4 \\

\tablemodellogo{\includegraphics[width=1.1em,height=0.95em,keepaspectratio]{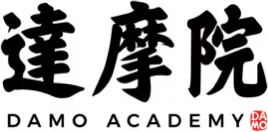}}VideoLLaMA2.1-7B-AV & 32 & 1.3/0.0 & 1.9/0.7 & 0.4 & 3.3/4.6 & 15.1/16.1 & 8.6 & 27.1/29.2 & 23.3 & 14.7/15.3 & 10.8 \\

\tablemodellogo{\includegraphics[width=1.1em,height=0.95em,keepaspectratio]{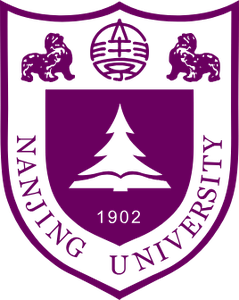}}VITA-1.5-7B & 16 & 0.0/0.8 & 0.5/1.2 & 0.2 & 0.0/0.0 & 1.1/1.3 & 0.5 & 34.2/34.6 & 27.5 & 11.9/12.4 & 9.4 \\
\bottomrule
\end{tabular}%
}
\vspace{-4mm}
\end{table*}\par\addvspace{0.5\baselineskip}\noindent\begin{minipage}{\linewidth}\setlength{\columnsep}{0.03\linewidth}\setlength{\intextsep}{4pt}\begin{wrapfigure}{r}{0.62\linewidth}\centering\setlength{\parskip}{0pt}\setlength{\abovecaptionskip}{5pt}\setlength{\belowcaptionskip}{0pt}\includegraphics[width=\linewidth]{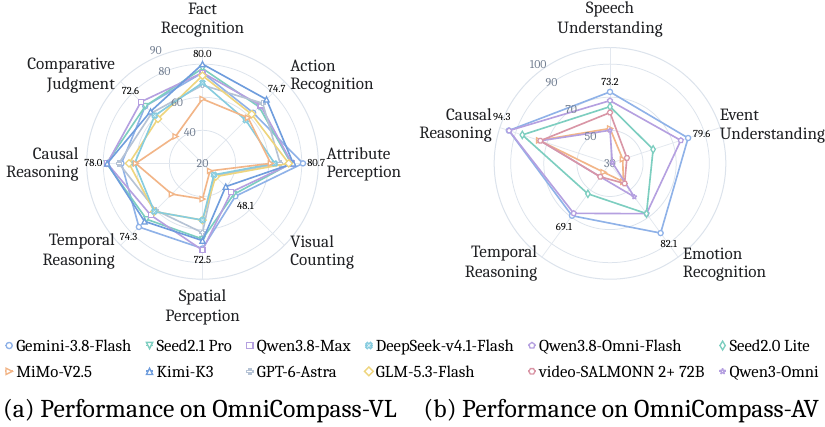}\caption{\textbf{Fine-grained QA performance across tracks.} We report category-wise $\mathrm{CS}^{\mathrm{QA}}$ scores in percent.}\label{fig:qa_performance}\end{wrapfigure}

\noindent\textbf{OmniCompass-AV.}
Table~\ref{tab:results_av} presents that Gemini-3.8-Flash leads in $\mathrm{CS}^{\mathrm{Avg}}$ at \textbf{57.4\%}, versus \textbf{24.6\%} for open-source leader MiMo-V2.5.
The results reflect the following insights.
\textbf{(i) Temporal Grounding and Reasoning Remain Challenging.}
Qwen3.8-Omni-Flash reaches 73.2\% in $\mathrm{CS}^{\mathrm{QA}}$ but only 21.8\% in $\mathrm{CS}^{\mathrm{TG}}$; all ten open-source models score at most \textbf{3.5\%} in $\mathrm{CS}^{\mathrm{TG}}$.
In Figure~\ref{fig:qa_performance}(b), Temporal Reasoning ranks lowest for five of six models, whereas Causal Reasoning leads all six at 74.3\%--94.3\%.
Qwen3-Omni-30B-A3B scores lowest on Event Understanding at 31.5\%.
\textbf{(ii) Open-Source Models Lag Most in Audio-Visual Event Binding.}
In Figure~\ref{fig:qa_performance}(b), the best proprietary and open-source $\mathrm{CS}^{\mathrm{QA}}$ scores differ by 12.5\% on Speech Understanding.
The gap reaches 38.9\% on Event Understanding, which requires linking audio cues to visible entities, actions, or scene context.
\textbf{(iii) Reference Effects and Consistency Gaps Persist with Audio.}
\textit{Text+Image} improves SG m\textsubscript{sIoU} and QA accuracy for 12 of 17 models each, but TG m\textsubscript{tIoU} for only five.
MiMo-V2.5 achieves 59.2\% and 58.7\% QA accuracy under \textit{Text} and \textit{Text+Image}, respectively, but only 47.3\% in $\mathrm{CS}^{\mathrm{QA}}$.
Similar aggregate accuracy therefore does not ensure correct answers on the same eligible pairs across settings.
\par\end{minipage}

Across both tracks, open-source models trail proprietary models, especially in audio-visual understanding.
Cross-reference consistency, fine-grained perception, and audio-visual event binding remain challenging.
Future models therefore need stronger multimodal alignment and audio-aware grounding.

\subsection{Analysis Experiments}
\label{subsec:analysis}

\begin{figure}[!t]
    \centering
    \setlength{\abovecaptionskip}{5pt}
    \setlength{\belowcaptionskip}{0pt}
    \includegraphics[width=\linewidth]{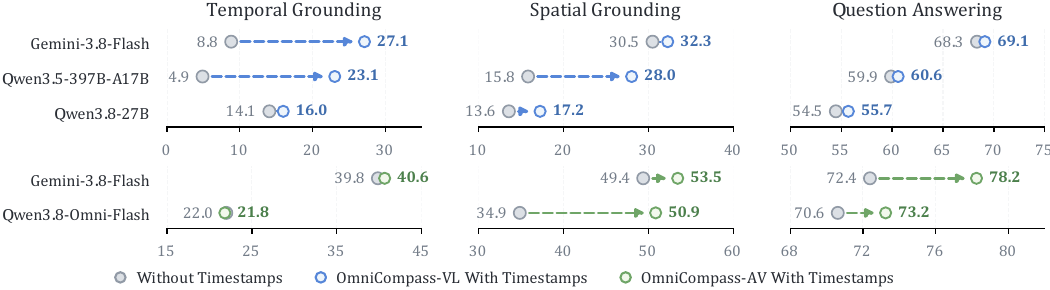}
    \vspace{-5mm}
    \caption{\textbf{Impact of frame-level timestamps.}
    Columns report $\mathrm{CS}^{\mathrm{TG}}$, $\mathrm{CS}^{\mathrm{SG}}$, and $\mathrm{CS}^{\mathrm{QA}}$ in percent.
    Dashed links connect scores without and with timestamps. Axis ranges vary by panel.}
    \vspace{-3mm}
    \label{fig:impact_frame_level_timestamps}\par\vspace{5mm}
    \centering
    \setlength{\abovecaptionskip}{5pt}
    \setlength{\belowcaptionskip}{0pt}
    \includegraphics[width=\linewidth]{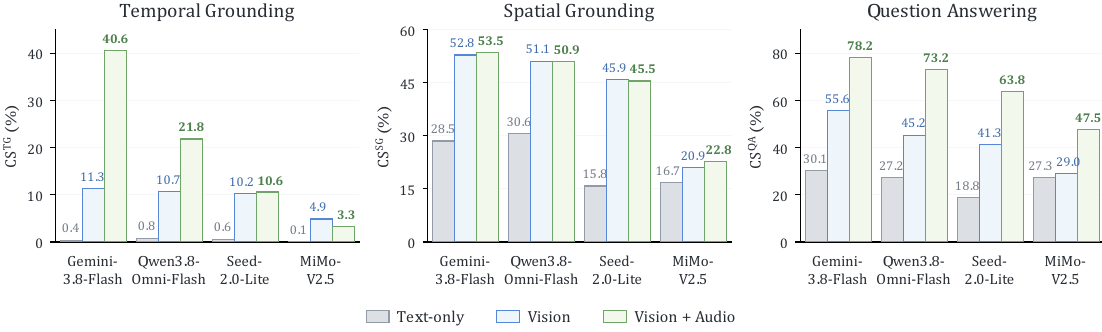}
    \vspace{-5mm}
    \caption{\textbf{Impact of input modalities on OmniCompass-AV.}
    Bars report CS in percent for Text-only, Vision, and Vision+Audio on matched eligible pairs within each model and task.}
    \vspace{-5mm}
    \label{fig:audio_gains}
\end{figure}

\noindent\textbf{Impact of Frame-Level Timestamps.}
Figure~\ref{fig:impact_frame_level_timestamps} shows that timestamps improve $\mathrm{CS}^{\mathrm{Avg}}$ in all five model-track configurations, with gains in \textbf{14 of 15} task-level comparisons.
On the VL track, Gemini-3.8-Flash, Qwen3.5-397B-A17B, and Qwen3.8-27B gain 7.0\%, 10.4\%, and 2.3\%, averaging \textbf{6.5\%}.
On the AV track, Gemini-3.8-Flash and Qwen3.8-Omni-Flash gain 3.6\% and 6.2\%, averaging \textbf{4.9\%}.
VL gains concentrate in TG: Gemini-3.8-Flash and Qwen3.5-397B-A17B gain \textbf{18.3\%} and \textbf{18.2\%}.
In AV, Qwen3.8-Omni-Flash gains \textbf{16.0\%} in SG, while TG changes by -0.2\% to +0.8\% across both models.
QA improves in every comparison by 0.7\%--5.9\%.
These results suggest that timestamps help align sampled frames with temporal positions.

\noindent\textbf{Impact of Input Modalities.}
Figure~\ref{fig:audio_gains} compares task-level CS across Text-only, Vision, and Vision+Audio for four OmniCompass-AV models.
All comparisons use matched eligible pairs within each model and task.
Text-only retains both question wordings but removes all media, including reference images.
Its $\mathrm{CS}^{\mathrm{TG}}$ is only 0.1--0.8\%, versus 18.8\%--30.1\% in $\mathrm{CS}^{\mathrm{QA}}$.
Vision improves all 12 model-task comparisons.
Adding audio raises $\mathrm{CS}^{\mathrm{QA}}$ by \textbf{18.6\%--28.1\%}, averaging \textbf{22.9\%}.
Gemini-3.8-Flash and Qwen3.8-Omni-Flash gain \textbf{29.4\%} and \textbf{11.1\%} in $\mathrm{CS}^{\mathrm{TG}}$, respectively, whereas MiMo-V2.5 loses 1.5\%.
$\mathrm{CS}^{\mathrm{SG}}$ changes range from -0.4\% to +1.8\%.
Thus, adding audio \textbf{consistently improves} $\mathrm{CS}^{\mathrm{QA}}$ across all four tested models, whereas TG gains vary by model and SG gains remain limited.
Additional analyses are provided in Appendix~\ref{sec:appendix_additional_analysis}.
\section{Conclusion}
In this paper, we introduce OmniCompass, a holistic benchmark for omni-modal video grounding and understanding.
It unifies TG, SG, and QA across VL and AV tracks with paired \textit{Text} and \textit{Text+Image} specifications for the same video, target, task semantics, and ground truth.
The benchmark covers 820 manually annotated videos across seven domains, including a 240-video AV subset with native audio.
The paired settings and CS metric assess both prediction correctness and consistency under changes in target specification.
Leading proprietary models outperform open-source models, especially in audio-visual understanding, while precise grounding, fine-grained perception, and audio-visual event binding remain challenging.
Reference images yield task-dependent effects, while timestamps improve model performance in both tracks.
These findings motivate stronger multimodal alignment and audio-aware grounding to connect target specifications with the relevant events and evidence.
OmniCompass supports joint evaluation of QA, precise grounding, and cross-reference consistency across tasks and modalities, providing a basis for diagnosing model limitations and measuring progress toward reliable omni-modal video understanding.
\bibliography{reference}
\bibliographystyle{iclr2027_conference}

\appendix

\counterwithin{table}{section}
\counterwithin{figure}{section}
\counterwithin{equation}{section}

\renewcommand{\thetable}{\Alph{section}\arabic{table}}
\renewcommand{\thefigure}{\Alph{section}\arabic{figure}}
\renewcommand{\theequation}{\Alph{section}\arabic{equation}}

\clearpage
\begingroup
\renewcommand{\topfraction}{0.95}
\renewcommand{\bottomfraction}{0.90}
\renewcommand{\textfraction}{0.05}
\renewcommand{\floatpagefraction}{0.75}
\section*{Appendix}
\pdfbookmark[0]{Appendix}{omnicompass-appendix}
\makeatletter
\let\ocSavedAddtocontents\addtocontents
\renewcommand{\addtocontents}[2]{%
  \def\ocCurrentExtension{#1}\def\ocMainContents{toc}%
  \ifx\ocCurrentExtension\ocMainContents
    \ocSavedAddtocontents{apx}{#2}%
  \else
    \ocSavedAddtocontents{#1}{#2}%
  \fi
}
\setcounter{tocdepth}{2}
\begingroup
\hypersetup{linkcolor=black}
\color{black}
\@starttoc{apx}
\endgroup
\makeatother
\medskip

\section{Additional Results}
\label{sec:appendix_additional_comparisons}

\subsection{Performance across Video Durations}
Figure~\ref{fig:vl_results_by_duration} compares ten models across short, medium, and long videos in OmniCompass-VL.
All ten score lower on long than short videos in TG.
Seed-2.1-Pro-260915 shows the largest decline, from 33.4\% to 15.1\%, while Gemini-3.8-Flash falls from 33.8\% to 21.0\%.
Only five models decline in SG and six in QA.
Thus, the direction of the duration association is more consistent for temporal localization than for the other tasks.
These groups also differ in content and target granularity, so the comparison does not isolate video duration as the cause of lower scores.

\begin{figure}[!htbp]
\centering
\includegraphics[width=\linewidth]{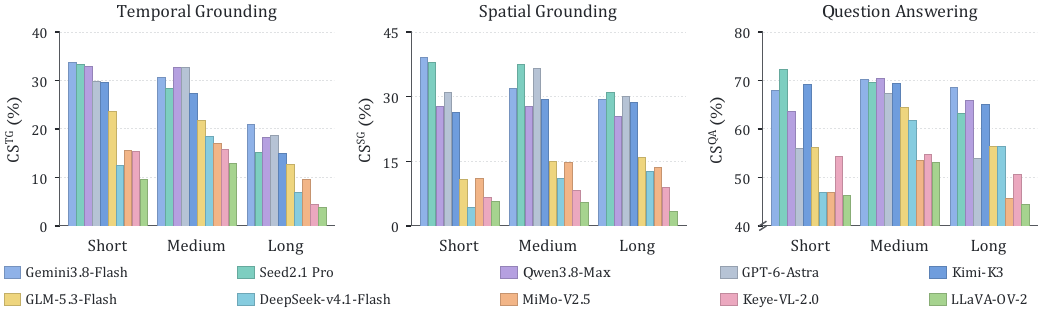}
\vspace{-5mm}
\caption{
\textbf{Performance across video durations in OmniCompass-VL.}
Bars show task-specific CS in percent for short videos below 90 seconds, medium videos from 90 to below 300 seconds, and long videos of at least 300 seconds.
Each model uses its own eligible pairs.
}
\vspace{-5mm}
\label{fig:vl_results_by_duration}
\end{figure}

\subsection{Grounding across Target Granularity}
Figure~\ref{fig:vl_grounding_granularity} compares ten VL models and six AV models across relative target-duration and box-area groups.
All ten VL models improve monotonically in TG as the relative target duration increases.
All six AV models also peak in the highest duration-ratio quartile.
Gemini-3.8-Flash rises from 7.9\% to 43.1\% on VL and from 27.1\% to 56.2\% on AV between the extreme groups.
Spatial trends differ: every VL model peaks in the largest-box group, whereas every AV model peaks in the third quartile.
Gemini-3.8-Flash scores 65.5\% in the third AV quartile and 53.6\% in the fourth.

Quartile boundaries are computed from all videos separately for each track; scores use the common eligible pairs within each task.
The resulting groups are not equal in size, and their boundaries differ between VL and AV.
IoU also depends on target extent, so these curves describe associations rather than an isolated granularity effect.
This contrast does not isolate audio effects: tasks, target distributions, and model rosters differ.

\begin{figure}[!htbp]
\centering
\includegraphics[width=\linewidth]{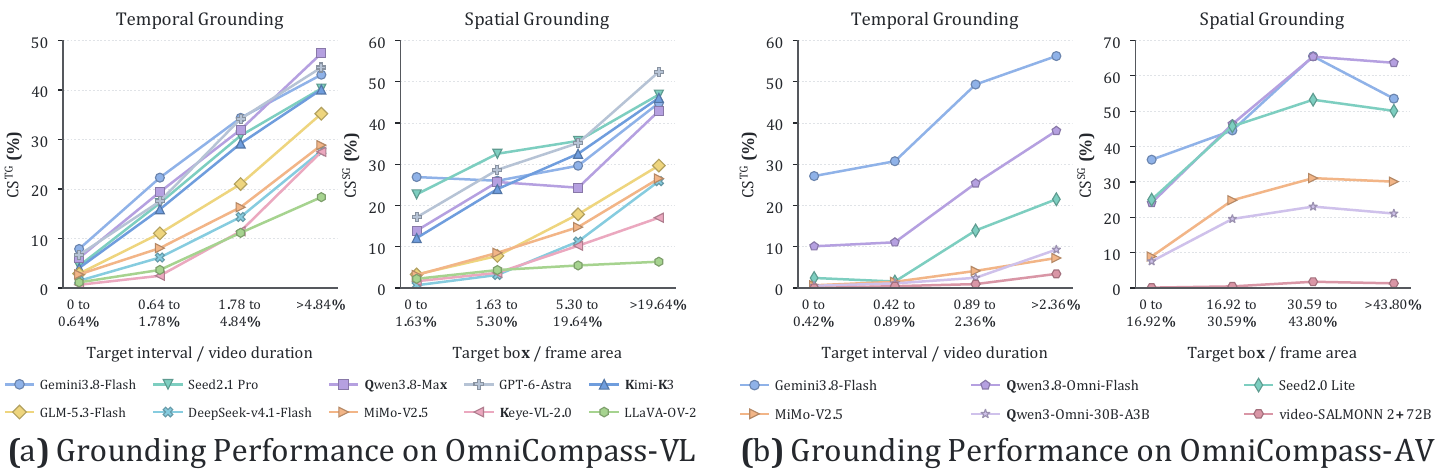}
\vspace{-5mm}
\caption{
\textbf{Grounding across target granularity.}
Curves report $\mathrm{CS}^{\mathrm{TG}}$ and $\mathrm{CS}^{\mathrm{SG}}$ in percent across quartiles of target-to-video duration and mean endpoint box-to-frame area.
Quartiles are track-specific; models share eligible pairs within each track and task.
}
\vspace{-5mm}
\label{fig:vl_grounding_granularity}
\end{figure}

\subsection{Performance across Video Domains}
Figure~\ref{fig:vl_domain_profiles} compares the ten VL models across seven domains.
All ten obtain their lowest SG and QA scores in Traffic, with equal-weight model means of 7.8\% and 37.3\%.
Variety has the lowest mean TG score at 8.1\%, but the highest mean SG score at 26.6\%.
Strong spatial grounding can therefore coexist with weak temporal localization within the same domain.
These rankings apply to the shared eligible subsets, not necessarily every video in each domain.
Domain comparisons also reflect differences in duration, target size, and QA content.

\begin{figure}[!htbp]
\centering
\includegraphics[width=\linewidth]{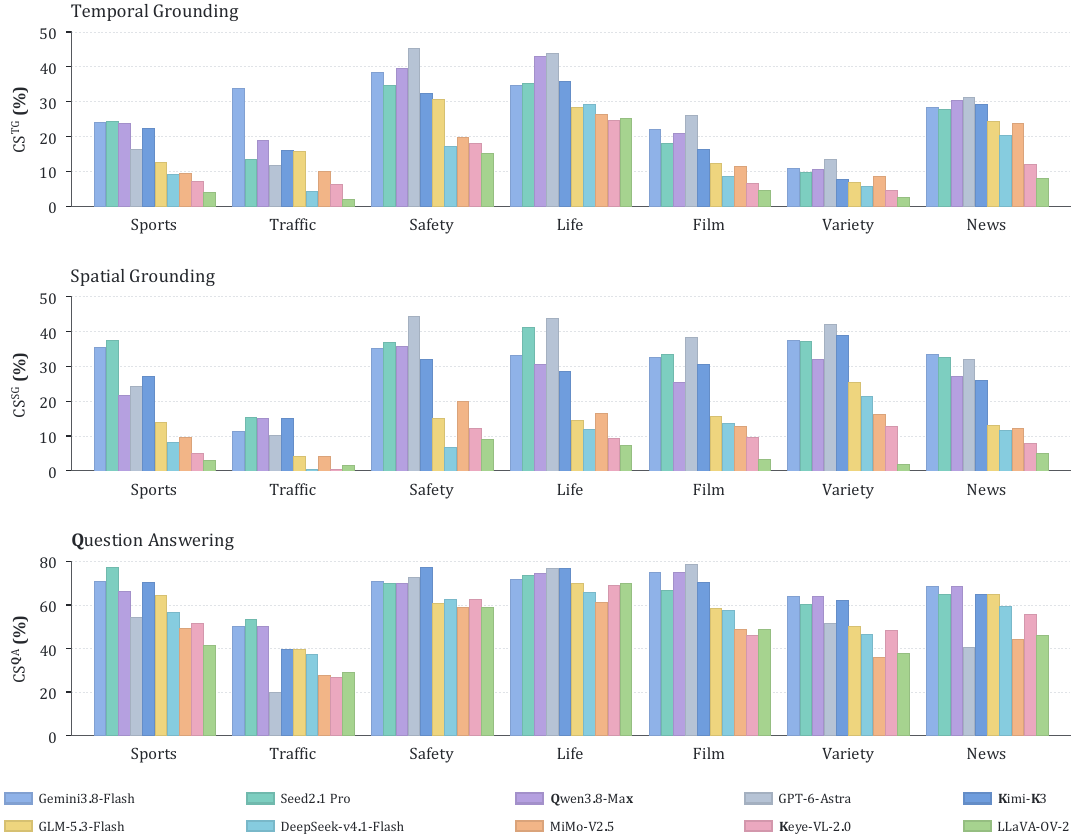}
\vspace{-5mm}
\caption{
\textbf{Performance across video domains in OmniCompass-VL.}
Bars report task-specific CS in percent for ten models.
The shared eligible subsets contain 592 TG, 673 SG, and 689 QA pairs across the seven domains; comparisons use the same instances within each task.
}
\vspace{-5mm}
\label{fig:vl_domain_profiles}
\end{figure}
\section{Additional Analyses}
\label{sec:appendix_additional_analysis}

\begin{figure*}[t]
\centering
\includegraphics[width=\textwidth]{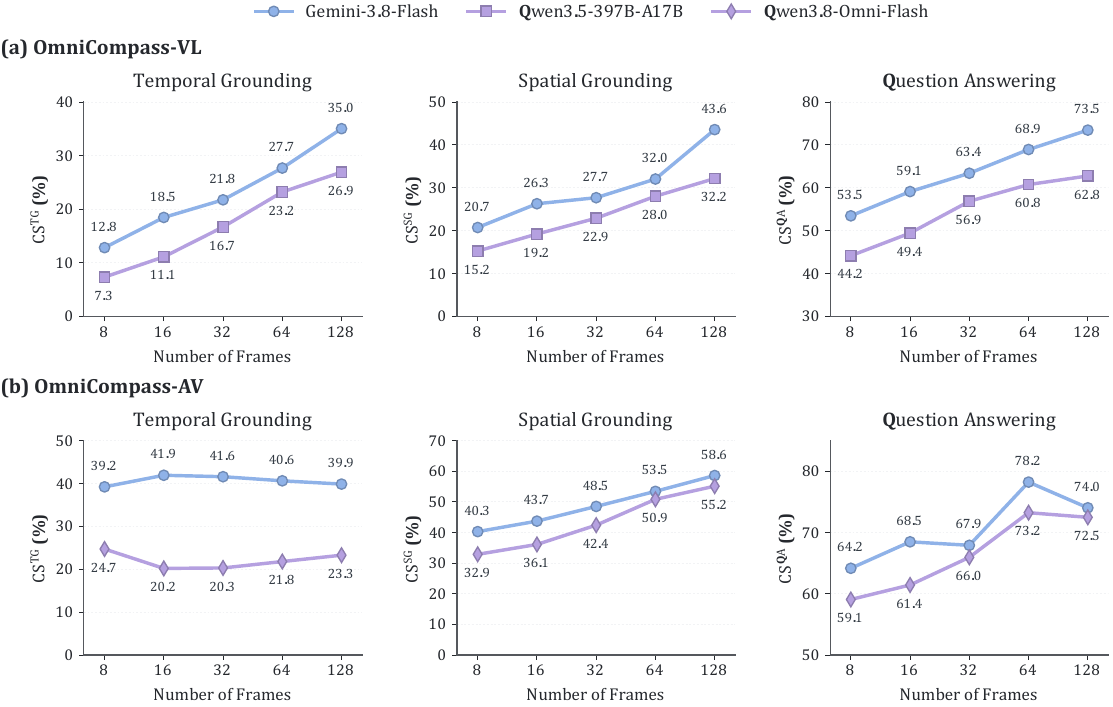}
\vspace{-5mm}
\caption{
\textbf{Impact of frame scaling.}
Each panel reports task-level paired CS (\%) under 8, 16, 32, 64, and 128 uniformly sampled frames.
Rows correspond to the VL and AV tracks, columns correspond to TG, SG, and QA, and curves distinguish the evaluated models.
}
\vspace{-2mm}
\label{fig:frame_scaling_vl_av}
\end{figure*}

\subsection{Impact of Frame Scaling}
Increasing the frame budget improves performance in most settings, but the effect depends on the track and task. On OmniCompass-VL, all three tasks improve significantly for both models from 8 to 128 frames. TG CS increases by \textbf{22.2\%} and 19.6\% for Gemini-3.8-Flash and Qwen3.5-397B-A17B, respectively; SG increases by \textbf{22.8\%} and 16.9\%, while QA increases by 20.0\% and 18.6\%. The gains are smaller and more task-dependent on OmniCompass-AV. SG improves by 18.3\% for Gemini-3.8-Flash and \textbf{22.3\%} for Qwen3.8-Omni-Flash, while QA improves by 9.9\% and 13.4\% from 8 to 128 frames. In contrast, TG remains nearly unchanged. We suppose that audio already provides useful temporal cues for AV TG (see Figure~\ref{fig:audio_gains}), leaving denser visual sampling with limited additional temporal evidence, whereas SG and QA can benefit from finer spatial details and richer visual context. Besides, more frames also increase visual input and inference cost, so the preferred budget depends on the task and track rather than always favoring the largest setting.

\subsection{Impact of Reference Specification}
Figure~\ref{fig:vl_reference_cue_effects} compares \textit{Text} and \textit{Text+Image} across reference types.
For action-conditioned questions, \textit{Text+Image} improves TG for seven of ten models and SG for nine, but lowers QA accuracy for eight.
Gemini-3.8-Flash gains 10.4\% in action-conditioned SG, whereas Qwen3.8-Max loses 7.6\%.
The paired settings change the target description as well as image availability, so these differences are not a pure image-addition effect.
They measure whether performance changes when a visual cue replaces part of the textual target description under matched evidence and semantics.

Each task uses the same eligible subset across all ten models.
Action-conditioned TG, SG, and QA contain 50, 54, and 31 pairs, respectively.
Scene-conditioned TG contains only eight pairs, so its cell values provide limited evidence for a general trend.
The heatmap reports mean differences, not significance; subgroup sizes matter when interpreting their direction.

\begin{figure}[!htbp]
\centering
\includegraphics[width=\linewidth]{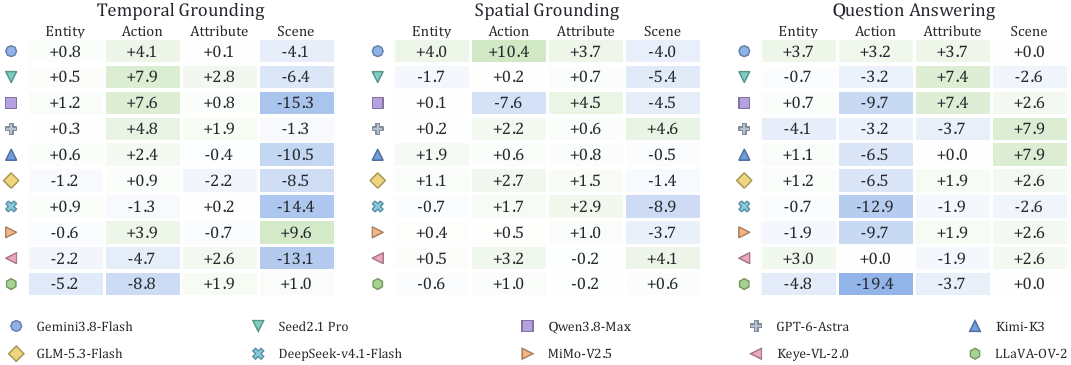}
\vspace{-5mm}
\caption{
\textbf{Effects of reference specification in OmniCompass-VL.}
Cells show \textit{Text+Image} minus \textit{Text} in percent for TG m\textsubscript{tIoU}, SG m\textsubscript{sIoU}, and QA accuracy on shared per-task subsets.
Green denotes increases and blue denotes decreases; color does not encode significance.
}
\vspace{-5mm}
\label{fig:vl_reference_cue_effects}
\end{figure}

\subsection{Cross-Specification Consistency}
Figure~\ref{fig:vl_reference_consistency} compares paired CS with setting-specific performance.
Across the ten models, TG and SG CS average 5.0\% and 5.7\% below the lower single-setting mean.
This gap includes both the cost of requiring overlap with ground truth under both settings and the additional prediction-to-prediction overlap constraint.
The latter contributes only 0.3\%--1.8\% across the model--task combinations.
The full gap does not measure prediction disagreement alone.

For QA, 10.5--25.1\% of eligible pairs are answered correctly under exactly one setting.
Kimi-K3 reaches nearly identical accuracies of 73.0\% and 72.9\%, yet answers only 67.6\% correctly under both.
Its 808 pairs comprise 546 correct under both settings, 44 only under \textit{Text}, 43 only under \textit{Text+Image}, and 175 under neither.
Similar aggregate accuracy can therefore mask different instance-level outcomes.
Neither-correct pairs may still contain two different wrong answers.

\begin{figure}[!htbp]
\centering
\includegraphics[width=\linewidth]{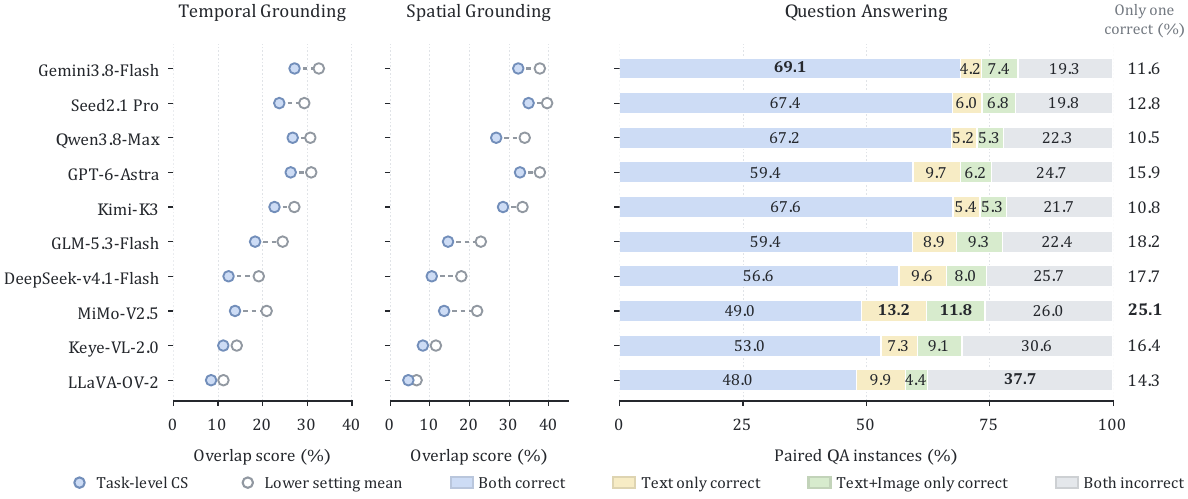}
\vspace{-5mm}
\caption{
\textbf{Consistency under paired reference specifications.}
Grounding compares CS with the lower setting-specific mean on identical eligible pairs.
QA bars partition pairs by correctness in the two settings; blue denotes $\mathrm{CS}^{\mathrm{QA}}$.
The right column reports correctness under exactly one setting.
Bold indicates category maxima, not uniformly better performance.
}
\vspace{-5mm}
\label{fig:vl_reference_consistency}
\end{figure}

\subsection{Qualitative Analysis}

\begin{figure}[!htbp]
\centering
\includegraphics[width=\linewidth]{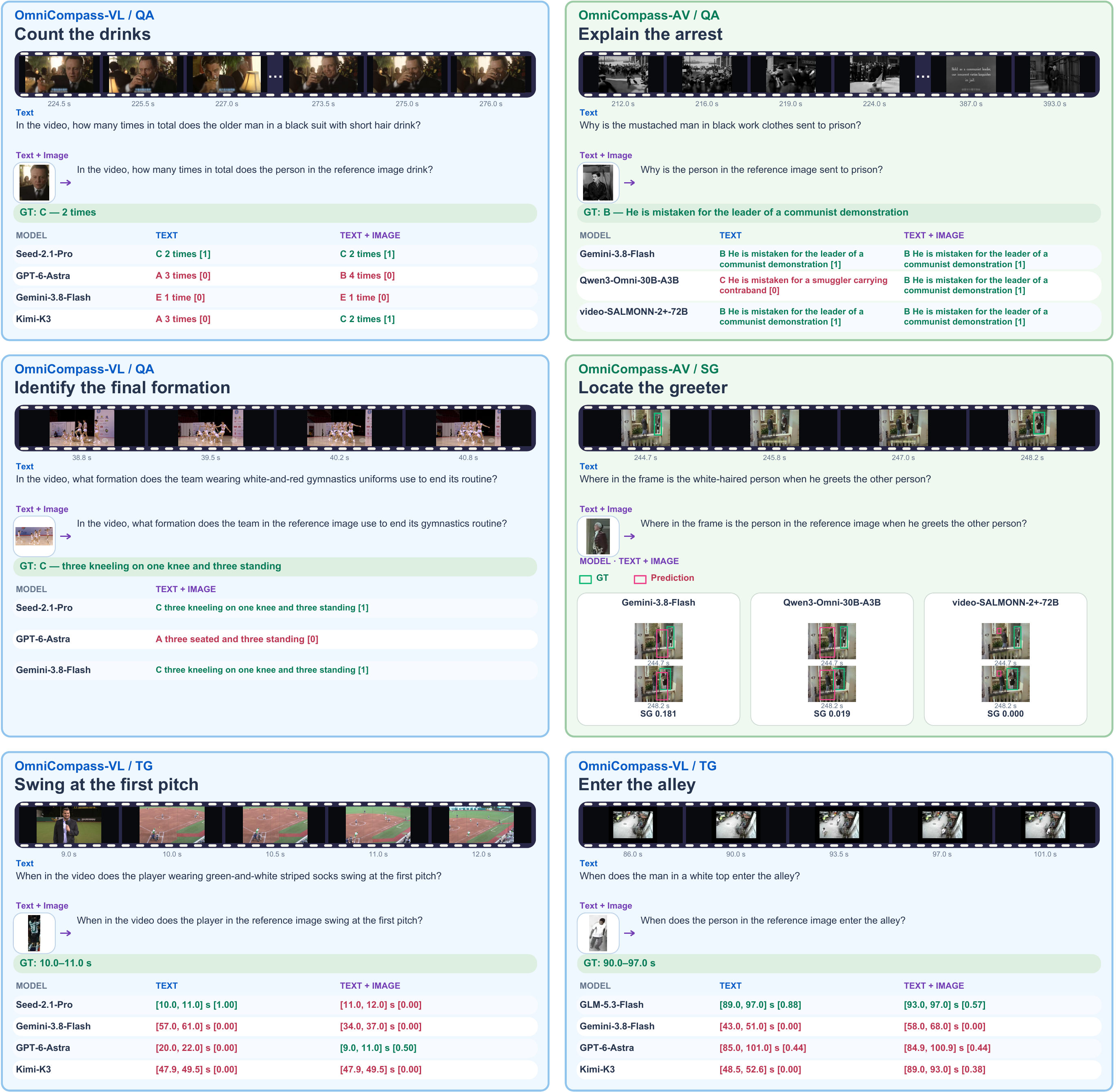}
\vspace{-5mm}
\caption{
\textbf{Qualitative Analysis on our OmniCompass-VL and OmniCompass-AV.}  Different models predictions on six cases in our two evaluation tracks.
}
\vspace{-5mm}
\label{fig:qualitative_analysis}
\end{figure}

Figure~\ref{fig:qualitative_analysis} shows qualitative examples of OmniCompass-VL and OmniCompass-AV across QA, SG, and TG. In the first OmniCompass-VL case, the older man drinks twice, corresponding to option C. Seed-2.1-Pro predicts C under both \textit{Text} and \textit{Text+Image}. Switching from \textit{Text} to \textit{Text+Image} changes Kimi-K3's prediction from A to C and GPT-6-Astra's from A to B. In the second OmniCompass-VL case, the gymnastics team ends with three members kneeling on one knee and three standing, corresponding to option C. Seed-2.1-Pro and Gemini-3.8-Flash select C, whereas GPT-6-Astra selects A.

For OmniCompass-AV, the upper-right QA example asks why the man is sent to prison. He is mistaken for the leader of a communist demonstration, corresponding to option B. Qwen3-Omni-30B-A3B changes from C under \textit{Text} to B under \textit{Text+Image}, while Gemini-3.8-Flash and video-SALMONN-2+-72B select B in both settings. The lower-right SG example illustrates the difficulty of SG: Gemini-3.8-Flash obtains the highest score among the three models at only 0.181. Qwen3-Omni-30B-A3B and video-SALMONN-2+-72B score 0.019 and 0.000, respectively.

\section{Benchmark Construction}
\label{sec:appendix_benchmark_construction}

\subsection{Video Sources and Selection}

\begin{figure}[!htbp]
\centering
\includegraphics[width=\linewidth]{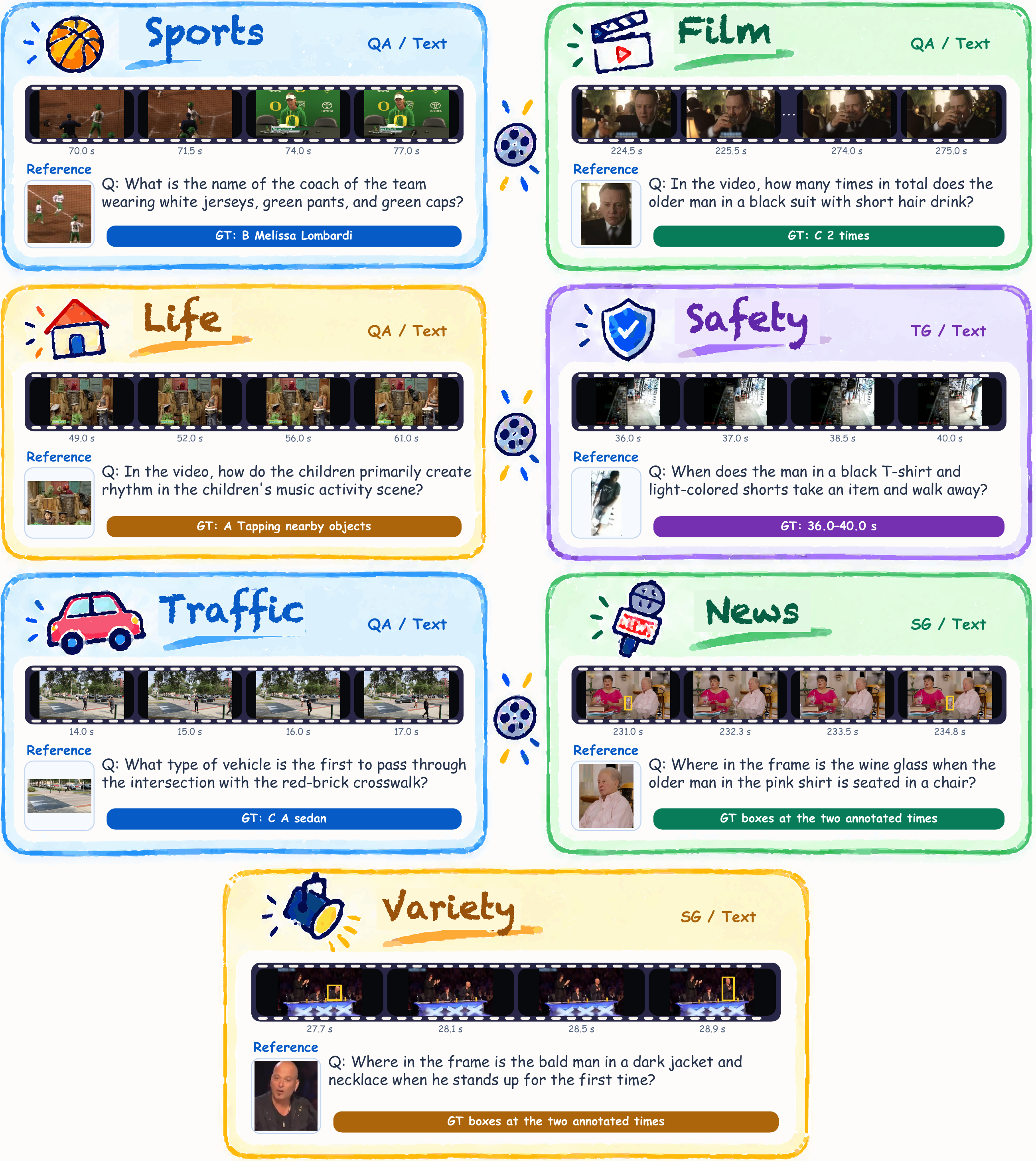}
\vspace{-5mm}
\caption{
\textbf{Seven high-level domains of videos in OmniCompass.}
}
\vspace{-3mm}
\label{fig:video_domains}
\end{figure}

OmniCompass-VL contains 580 selected public-source videos and 240 self-collected clips.
The public component draws from nine dataset families; Table~\ref{tab:source_coverage} reports the contribution of each source to the 820-video set.
The self-collected component comprises 70 movie clips, 70 TV-series clips, 50 variety-show clips, and 50 news clips.
These same 240 videos also support OmniCompass-AV, with separate audio-visual task annotations.
The two tracks share videos, but their TG, SG, and QA annotations may select different events and targets from that shared video content. Figure ~\ref{fig:video_domains} summarizes the seven high-level video categories in OmniCompass.

\begin{table*}[!htbp]
\centering
\caption{
\textbf{Nine public source datasets contributing to OmniCompass.}
Counts refer to selected videos, not complete source releases.
Domains follow the final OmniCompass taxonomy; News includes documentary, nature, and knowledge content.
Links point to downloads.
}
\label{tab:source_coverage}
\footnotesize
\setlength{\tabcolsep}{3pt}
\renewcommand{\arraystretch}{1.12}
\begin{threeparttable}
\begin{tabular}{@{}>{\raggedright\arraybackslash}p{0.235\linewidth}r>{\raggedright\arraybackslash}p{0.13\linewidth}>{\raggedright\arraybackslash}p{0.17\linewidth}>{\raggedright\arraybackslash}p{0.338\linewidth}@{}}
\toprule
\textbf{Source} & \textbf{Videos} & \textbf{Selected domains} & \textbf{License / Terms} & \textbf{Link} \\
\midrule
\texttt{MultiSports}\newline\citep{li2021multisports} & 83 & Sports & CC BY-NC 4.0 & {\scriptsize\url{https://huggingface.co/datasets/MCG-NJU/SportsAction}} \\
\texttt{LiveSports-3K}\newline\citep{chen2025livecc} & 92 & Sports & Not specified & {\scriptsize\url{https://huggingface.co/datasets/stdKonjac/LiveSports-3K}} \\
\texttt{UCF-Crime}\newline\citep{sultani2018UCF-Crime} & 111 & Safety & CC0-1.0 on mirror & {\scriptsize\url{https://huggingface.co/datasets/AllenXeon/ucf_crime}} \\
\texttt{CG-Bench}\newline\citep{chen2025cgbench} & 58 & Film, Variety, News, Life, Safety, Traffic & MIT; videos for academic research only & {\scriptsize\url{https://huggingface.co/datasets/CG-Bench/CG-Bench}} \\
\texttt{MovieChat-1K-test}\newline\citep{song2024moviechat} & 20 & News & Not specified & {\scriptsize\url{https://huggingface.co/datasets/Enxin/MovieChat-1K-test}} \\
\texttt{ActivityNetQA}\newline\citep{yu2019activitynet} & 37 & Life & MIT; non-commercial use & {\scriptsize\url{https://huggingface.co/datasets/lmms-lab/ActivityNetQA}} \\
\texttt{QVHighlights}\newline\citep{lei2021QVHighlights} & 88 & Life, News & CC BY-NC-SA 4.0 for annotations & {\scriptsize\url{https://huggingface.co/datasets/yeliudev/VideoMind-Dataset/tree/main/qvhighlights}} \\
\texttt{StreetScene}\newline\citep{ramachandra2020streetscene} & 32 & Traffic & CC BY-SA 4.0 & {\scriptsize\url{https://zenodo.org/records/10870472}} \\
\texttt{ComplexVAD}\newline\citep{mumcu2025complexvad} & 59 & Traffic & CC BY-SA 4.0 & {\scriptsize\url{https://zenodo.org/records/15707073}} \\
\bottomrule
\end{tabular}
\end{threeparttable}
\end{table*}

Table~\ref{tab:source_coverage} documents provenance rather than a blanket license for the underlying media.
Our annotations and the source videos are distinct assets; reuse of a dataset repository does not itself establish permission to redistribute its videos.
The self-collected clips are used for academic evaluation, with rights to the original programs retained by their owners.
Access and redistribution must respect the applicable source terms, including any restrictions on the videos and reference images.

We use \texttt{Qwen3-VL-8B-Instruct}~\citep{Qwen3-VL} to profile 8,489 domain-aligned public candidates.
The outputs describe entities, actions, attributes, and scenes, together with a video description and corresponding counts.
The archived records use \texttt{identity} for entity information.
Figure~\ref{fig:qwen_structured_prompt} presents the profiling template with this field displayed as \texttt{entity} for consistency with the paper's terminology.
These automatic descriptions guide candidate selection; annotators construct questions, localize the required evidence, and verify answers.

\begin{figure*}[h]
\centering
\begin{omniprompt}{Structured Video Profiling Prompt}
\small
You are a factual video captioning and structured extraction assistant.\\
Only describe what is clearly visible or audible. Do NOT guess.\\
Output ONE valid JSON object only.\\
Do NOT place JSON inside any field.
\smallskip
\textbf{Write:}
\begin{enumerate}[leftmargin=18pt, topsep=2pt, itemsep=0pt, parsep=0pt]
\item \texttt{detailed\_description}: 3--6 sentences, time order.
\item \texttt{entity}: list categories with \texttt{instance\_count} and short notes.
\item \texttt{action}: up to 12 distinct temporally localizable action variants.
\item \texttt{attribute}: up to 12 distinct visible cues.
\item \texttt{scene}: up to 6 distinct phases, shots, or contexts.
\item \texttt{counts}: \texttt{entity} = sum of \texttt{entity.instance\_count}; others = list lengths.
\end{enumerate}
{\normalfont\bfseries Required JSON Output\par}
\smallskip
\footnotesize\ttfamily
Return JSON with EXACT keys:\\
\{\\
"detailed\_description":"...",\\
"entity":[\{"category":"...","instance\_count":0,"notes":"..."\}],\\
"action":["..."],\\
"attribute":["..."],\\
"scene":["..."],\\
"counts":\{"entity":0,"action":0,"attribute":0,"scene":0\}\\
\}
\end{omniprompt}
\caption{\textbf{Structured video profiling prompt used with \texttt{Qwen3-VL-8B-Instruct}.}}
\label{fig:qwen_structured_prompt}
\end{figure*}

Deterministic filtering removes 1,454 candidates shorter than 30 seconds, leaving 7,035 videos.
A further 89 lack at least one of the four cue types, yielding 6,946 eligible public candidates.
The filtering criteria require positive entity, action, attribute, and scene counts alongside the metadata used by the selection pipeline.
These counts describe the candidate pool before manual task construction, not validated benchmark questions.

\subsection{Task-Oriented Candidate Scoring}
Candidate ranking combines reference necessity $R$, grounding suitability $G$, temporal-spatial utility $T$, and evidence diversity $E$ as
$S=0.35R+0.25G+0.25T+0.15E$.
These heuristic components use structured profiling statistics, not model evaluation scores.
Table~\ref{tab:qwen_scoring} defines the recorded component calculations on a zero-to-ten scale.
Scores inform shortlisting together with domain, duration, source, and subcategory coverage; human curation determines the retained public set.

\begin{table*}[t]
\caption{
\textbf{Mapping from structured video profiling outputs to task-oriented scoring signals.}
We summarize how the structured video profiling outputs produced by \texttt{Qwen3-VL-8B-Instruct} are converted into intermediate statistics and then used to compute the four score components.
}
\label{tab:qwen_scoring}
\centering
\setlength{\tabcolsep}{4.2pt}
\renewcommand{\arraystretch}{1.08}
\resizebox{\textwidth}{!}{
\begin{threeparttable}
\begin{tabular}{p{2.5cm} p{3.8cm} p{6.1cm} p{4.8cm}}
\toprule
\textbf{Component} & \textbf{Derived profiling fields} & \textbf{Computation} & \textbf{Intuition} \\
\midrule

\textbf{Reference Necessity ($R$)}
& \texttt{entity\_count}, \texttt{action\_count}, \texttt{attribute\_count}, \texttt{scene\_count}, repeated actions
& We first compute \textit{ambiguity} and \textit{richness}. Ambiguity increases with multiple candidate entities, repeated actions, scene variation, and potential attribute confusion. Richness increases with broader coverage over entity, action, attribute, and scene signals. We then combine them as
$
R = 10 \cdot \left(0.60A + 0.20B + 0.20AB\right),
$
where $A$ and $B$ are the normalized ambiguity and richness scores.
& Measures whether external referential cues are needed to uniquely specify the target. \\

\textbf{Grounding Suitability ($G$)}
& \texttt{entity\_count}, \texttt{action\_count}, \texttt{attribute\_count}, \texttt{scene\_count}
& We estimate temporal grounding potential from action and scene counts, spatial grounding potential from entity and attribute counts, and question answering potential from action and scene-related complexity. The final score is
$
G = 10 \cdot \mathrm{avg}(P_{\text{TG}}, P_{\text{SG}}, P_{\text{QA}}).
$
& Measures whether the video can support constructing TG, SG, and VideoQA instances. \\

\textbf{Temporal-Spatial Utility ($T$)}
& video duration, \texttt{entity\_count}, \texttt{action\_count}, \texttt{scene\_count}
& We combine a duration term and a variation term:
$
T = 10 \cdot \left(0.60D + 0.40V\right),
$
where $D$ is a log-normalized duration score and $V$ is the average of normalized action, scene, and entity variation.
& Measures whether the video provides sufficient temporal span and visual variation for evidence localization. \\

\textbf{Evidence Diversity ($E$)}
& \texttt{entity\_count}, \texttt{action\_count}, \texttt{attribute\_count}, \texttt{scene\_count}
& We compute normalized entropy over the four evidence counts:
$
E = 10 \cdot H_{\text{norm}}(c_{\text{ent}}, c_{\text{act}}, c_{\text{att}}, c_{\text{scn}}).
$
& Measures whether referential evidence is distributed across multiple cue types rather than concentrated in a single source. \\

\midrule
\textbf{Final Score ($S$)}
& $R, G, T, E$
& $
S = 0.35R + 0.25G + 0.25T + 0.15E
$
& Encourages videos that require diverse referential cues, support TG/SG/VideoQA construction, and provide diverse evidence for annotation. \\

\bottomrule
\end{tabular}
\vspace{-4mm}
\end{threeparttable}
}
\end{table*}

\subsection{Question-Answering Taxonomy}

\begin{figure}[!htbp]
\centering
\includegraphics[width=\linewidth]{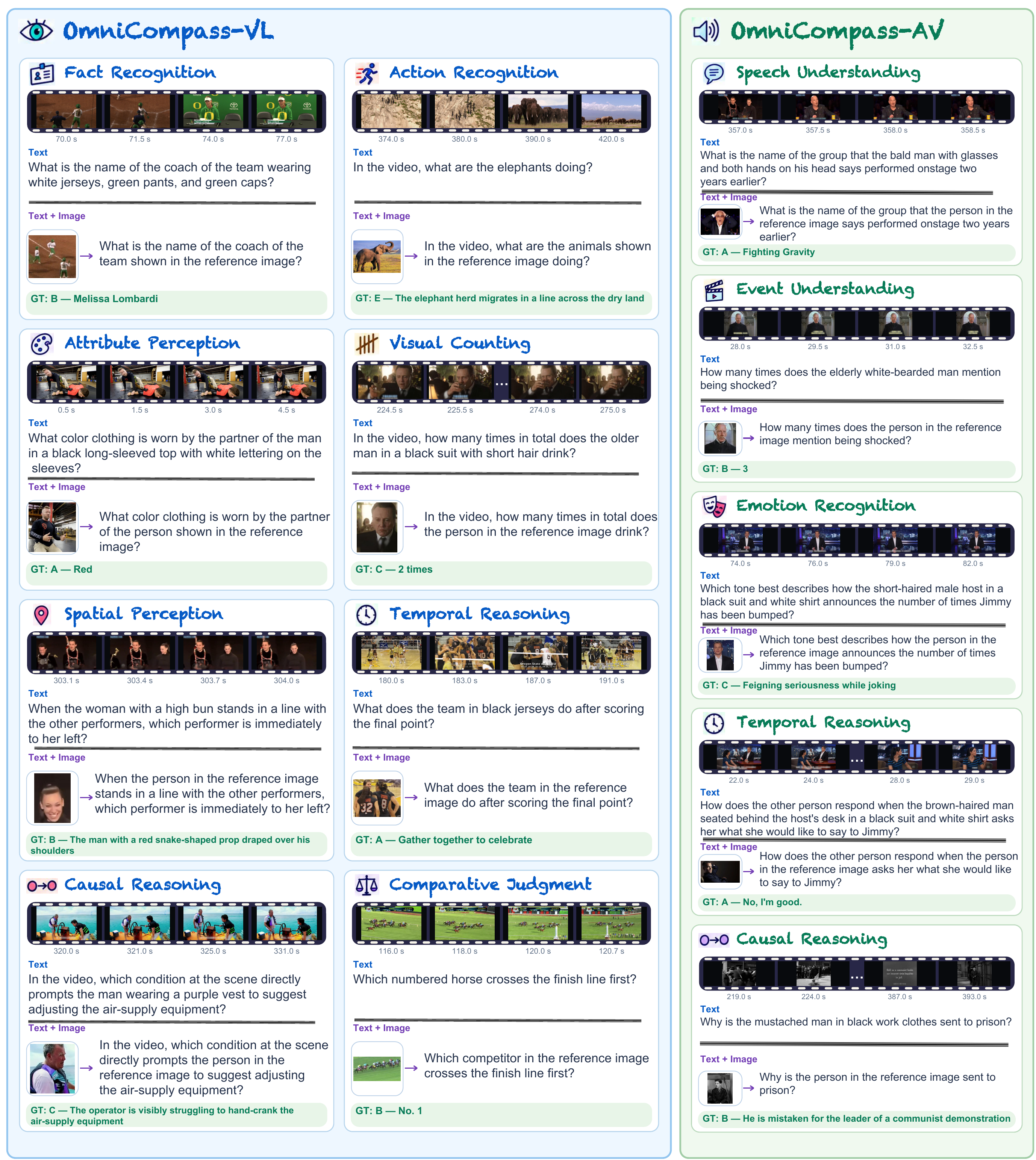}
\vspace{-5mm}
\caption{
\textbf{QA taxonomy in our OmniCompass-VL and OmniCompass-AV.}
}
\vspace{-5mm}
\label{fig:vl_av_13_class}
\end{figure}

QA types are grouped by the dominant operation needed to answer each question.
Perception covers direct recognition and judgments from the available evidence, while reasoning covers temporal, causal, or comparative relations.
Both tracks use five answer options, but their type sets differ because AV questions are designed to draw on speech, sound, or joint audio-visual evidence.
The shared format standardizes scoring without equating difficulty across tracks.

OmniCompass-VL contains eight QA types, listed in Table~\ref{tab:qa_taxonomy} and shown in Figure~\ref{fig:vl_av_13_class}.
Fact Recognition, Action Recognition, Attribute Perception, Visual Counting, and Spatial Perception form the perception group.
Temporal Reasoning, Causal Reasoning, and Comparative Judgment form the reasoning group.
Labels follow the answering operation, not individual words in the question.

\begin{table}[h]
\caption{\textbf{QA taxonomy and examples in OmniCompass-VL.} We group the eight QA types into Perception and Reasoning. Examples show only the question text from the \textit{Text} target-specification setting; answer options are omitted.}
\vspace{5pt}
\label{tab:qa_taxonomy}
\centering
\footnotesize
\setlength{\tabcolsep}{3pt}
\renewcommand{\arraystretch}{1.15}
\begin{threeparttable}
\begin{tabular}{@{}>{\raggedright\arraybackslash}p{.13\textwidth}>{\raggedright\arraybackslash}p{.22\textwidth}>{\raggedright\arraybackslash}p{.28\textwidth}>{\raggedright\arraybackslash}p{\dimexpr.37\textwidth-6\tabcolsep\relax}@{}}
\toprule
\textbf{Category} & \textbf{QA Task} & \textbf{Task Definition} & \textbf{Example} \\
\midrule
\textbf{Perception} & Fact Recognition & Identify entities or explicit facts associated with the referenced target. & In the video, when the man in a black suit sitting on the left side of the table is talking to two other people, what is the drink in front of the woman? \\
& Action Recognition & Identify an action performed by the referenced target. & In the video, what farewell gesture did the young man in the pink hoodie make to each of the two women inside the house before leaving? \\
& Attribute Perception & Identify the target's appearance, attributes, or observable state. & In the video, when a lady wearing a red-brimmed straw hat dances with a group of people in yellow clothes, what is the color of the oil-paper umbrella held by the lady behind her? \\
& Visual Counting & Count referenced entities, actions, or event occurrences. & In the video, how many rotations did the athlete complete while raising one hand and high-fiving a partner? \\
& Spatial Perception & Determine scene context, locations, or spatial relations involving the referenced target. & In the video, where is the kettle relative to the man in a gray hat and light blue short-sleeved shirt, based on the direction he is facing while making tea? \\
\midrule
\textbf{Reasoning} & Temporal Reasoning & Identify events or states before, during, or after a specific referenced event. & Before the man in a blue shirt enters the first elevator, what type of passenger is among the previous group entering the same elevator? \\
& Causal Reasoning & Infer causes, effects, or motives that connect observed events. & In the video, why does the boy with short blond hair repeatedly throw away the eggs in secret rather than return or refuse them directly? \\
& Comparative Judgment & Compare targets, events, states, or outcomes to determine differences, rankings, or changes. & In the video, when several people on stage take turns answering the question asked by the female host wearing a colorful patterned sleeveless top, what is the relationship among their answering durations? \\
\bottomrule
\end{tabular}
\end{threeparttable}
\end{table}

OmniCompass-AV contains five QA types, listed in Table~\ref{tab:qa_taxonomy_av} and shown in Figure~\ref{fig:vl_av_13_class}.
Speech Understanding, Event Understanding, and Emotion Recognition form the perception group; Temporal Reasoning and Causal Reasoning form the reasoning group.
Event Understanding links audible cues to visible participants, actions, states, or context.
These labels describe intended evidence requirements, rather than a measured guarantee that every item fails when audio is removed.

\begin{table}[h]
\caption{\textbf{QA taxonomy and examples in OmniCompass-AV.} We group the five QA types into Perception and Reasoning. Examples show only the question text from the \textit{Text} target-specification setting; answer options are omitted.}
\vspace{5pt}
\label{tab:qa_taxonomy_av}
\centering
\footnotesize
\setlength{\tabcolsep}{3pt}
\renewcommand{\arraystretch}{1.15}
\begin{threeparttable}
\begin{tabular}{@{}>{\raggedright\arraybackslash}p{.13\textwidth}>{\raggedright\arraybackslash}p{.22\textwidth}>{\raggedright\arraybackslash}p{.28\textwidth}>{\raggedright\arraybackslash}p{\dimexpr.37\textwidth-6\tabcolsep\relax}@{}}
\toprule
\textbf{Category} & \textbf{QA Task} & \textbf{Task Definition} & \textbf{Example} \\
\midrule
\textbf{Perception} & Speech Understanding & Understand explicit speech content, including facts, quantities, descriptions, and opinions. & How does the gray-haired man in a black leather jacket and white top summarize what the Nike swoosh represented to young people at the time? \\
& Event Understanding & Identify participants, objects, actions, states, or counts in audio-visual events. & When the woman with tied-up hair, a blue off-the-shoulder top, and brown overalls introduces Sean's recently deceased father, which pair of objects appears beside the relevant framed photograph? \\
& Emotion Recognition & Recognize emotion, tone, attitude, or affective reactions from audio-visual evidence. & In the video, after repeated corrections from the man in red, how does the woman with shoulder-length auburn hair sound when asking whether he wants to tell the story? \\
\midrule
\textbf{Reasoning} & Temporal Reasoning & Determine the order or temporal relations of utterances, events, and states. & In the video, before the blonde woman calls the grey bearded man arrogant, what is he doing? \\
& Causal Reasoning & Infer causes, motives, or consequences from audio-visual context. & Why could the lie told by the woman in a red choker and black top to avoid awkwardness with Joey end up costing her job? \\
\bottomrule
\end{tabular}
\end{threeparttable}
\end{table}

\subsection{Reference Design and Provenance}
Reference images identify a target through entity, action, attribute, or scene information, as defined in Table~\ref{tab:ref_types}.
The unit of analysis is a task-specific reference, not a unique image file or video.
Each of the 820 VL videos contributes one reference for TG, SG, and QA, yielding 2,460 task instances.
Tasks can select different targets; their cue types and sources are counted separately.

\begin{table}[h]
\caption{\textbf{Functional taxonomy of reference information in OmniCompass-VL.}
Each type states what the reference image contributes to target identification and gives representative examples.}
\label{tab:ref_types}
\centering
\small
\setlength{\tabcolsep}{6pt}
\renewcommand{\arraystretch}{1.12}
\begin{threeparttable}
\begin{tabular}{p{0.16\textwidth}p{0.78\textwidth}}
\toprule
\textbf{Reference Type} & \textbf{Role in Target Identification and Typical Examples} \\
\midrule
\texttt{Entity} & Shows a particular person or object that the model must match across views, such as an athlete, actor, vehicle, or bag. \\
\texttt{Action} & Shows a recognizable motion or event pattern that defines the target behavior, such as a dunk, basketball shot, header, or tackle. \\
\texttt{Attribute} & Distinguishes the target through a visible property, appearance, or state, such as clothing color, vehicle type, shape, or pose. \\
\texttt{Scene} & Shows a place or surrounding visual context to match with a video segment, such as a street, room, stage, or stadium. \\
\bottomrule
\end{tabular}
\end{threeparttable}
\end{table}

\begin{figure}[!htbp]
\centering
\includegraphics[width=\linewidth]{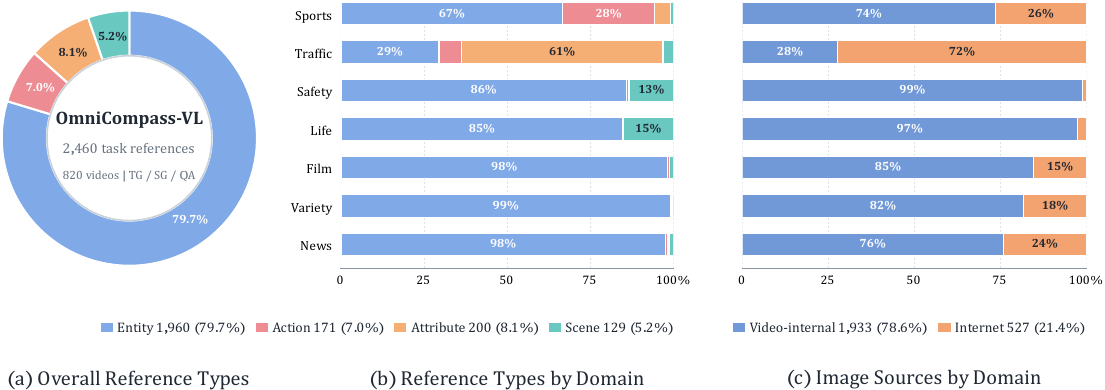}
\vspace{-5mm}
\caption{
\textbf{Reference composition across 2,460 VL task instances.}
Panel~\textbf{(a)} shows overall cue-type frequencies, \textbf{(b)} shows their proportions by domain, and \textbf{(c)} compares video-internal and Internet sources.
Counts are reported per task instance, so an image reused across tasks is counted more than once.
}
\vspace{-5mm}
\label{fig:reference_type_source_composition}
\end{figure}

Entity cues account for 1,960 of the 2,460 references, or 79.7\%.
The remaining references comprise 171 action cues, 200 attribute cues, and 129 scene cues.
The examples include recognizable actions in Sports and appearance-based vehicle distinctions in Traffic.
Figure~\ref{fig:reference_type_source_composition} reports domain-wise frequencies; the cue types are not equally represented in this set.

The reference set contains 1,933 video-internal images and 527 Internet images.
Video-internal crops preserve the identity of targets specific to a recording, while external exemplars can specify recognizable people, objects, or actions.
Source selection follows the identification needs of each annotation rather than a fixed per-domain quota.
This variation is part of the benchmark's reference conditions and is not an independently controlled source-modality experiment.

For video-internal references, annotation guidelines favor a separate view outside the target interval or QA clue interval whenever one is available.
Reviewers also inspect whether the image reveals the answer event, since temporal separation alone does not rule out semantic leakage.
Crops retain target-identifying content while limiting irrelevant context and direct answer evidence.
These checks aim to separate target specification from the evidence needed to solve the task.
\section{Annotation Quality and Verification}
\label{sec:appendix_quality_verification}

\subsection{Review Criteria and Repair Records}
Quality control combines semantic review with structural validation.
Reviewers check paired-setting target identity, temporal-event completeness, endpoint box alignment, QA answer uniqueness, and evidence sufficiency.
AV review additionally examines speech or sound dependence and audio-visual correspondence.
Returned corrections undergo task-level re-verification and lead-expert acceptance.
Tables~\ref{tab:annotation_quality_control_vl} and~\ref{tab:annotation_quality_control_av} summarize verification stages and representative issues.

\subsection{Human Effort}
The construction and verification records yield a retrospective estimate of 1,784 human-hours, or 223 eight-hour person-days.
This comprises 1,280 annotator-hours and 504 lead-expert-hours, as detailed in Table~\ref{tab:annotation_effort}.
The estimate covers training, task construction, cross-verification, correction, adjudication, and final auditing.
It is reconstructed from project records rather than continuous time logs and excludes automated processing, model inference, and waiting periods.

\begingroup
\let\ocQualityTable\table
\let\ocQualityEndTable\endtable
\renewenvironment{table*}[1][]{\ocQualityTable[!htbp]}{\ocQualityEndTable}
\BeforeBeginEnvironment{tabular}{\begin{adjustbox}{max width=\linewidth}}
\AfterEndEnvironment{tabular}{\end{adjustbox}}
\begin{table*}[t]
\centering
\caption{
\textbf{Multi-stage quality verification for OmniCompass-VL.}
All counts use the final 820-video release set. A video passes only when its paired \textit{Text} and \textit{Text+Image} specifications and its TG, SG, and QA annotations satisfy the verification criteria. Representative findings are non-exclusive.
}
\label{tab:annotation_quality_control_vl}
\footnotesize
\setlength{\tabcolsep}{3.5pt}
\renewcommand{\arraystretch}{1.12}
\begin{tabular}{p{0.20\textwidth} p{0.08\textwidth} p{0.20\textwidth} p{0.25\textwidth} p{0.17\textwidth}}
\toprule
\textbf{Verification stage} & \textbf{Videos} & \textbf{Stage outcome} & \textbf{Representative findings} & \textbf{Cumulative pass} \\
\midrule
Annotator cross-verification
& 820
& 466 require no revision; 354 are returned for revision.
& Cross-setting target mismatch; a \textit{Text+Image} pointer identifiable without its image; imprecise TG boundaries; incorrect or incomplete SG targets; insufficient QA evidence, non-unique answers, or weak distractors.
& \textbf{466/820} (\textbf{56.8\%}) \\
\midrule
Annotator revision and lead-expert acceptance
& 354
& All revised cases pass after annotator recheck and lead-expert acceptance.
& Typical repairs address overly small or partial SG targets, TG intervals that omit part of the event, and inconsistencies among QA questions, answers, types, and evidence.
& \textbf{820/820} (\textbf{100\%}) \\
\midrule
Final release audit
& 820
& All VL cases are included only after lead-expert acceptance.
& Same target and ground truth across settings; complete TG events; valid boundary-frame boxes; one evidence-supported QA answer with plausible distractors.
& \textbf{820/820} (\textbf{100\%}) \\
\bottomrule
\end{tabular}
\end{table*}

\begin{table*}[t]
\centering
\caption{
\textbf{Multi-stage quality verification for OmniCompass-AV.}
All counts use the 240-video AV subset of the final release. In addition to the shared VL criteria, annotators check native-audio dependence and audio-visual alignment, and the lead expert includes these criteria in final acceptance and audit. Representative findings are non-exclusive.
}
\label{tab:annotation_quality_control_av}
\footnotesize
\setlength{\tabcolsep}{3.5pt}
\renewcommand{\arraystretch}{1.12}
\begin{tabular}{p{0.20\textwidth} p{0.08\textwidth} p{0.20\textwidth} p{0.25\textwidth} p{0.17\textwidth}}
\toprule
\textbf{Verification stage} & \textbf{Videos} & \textbf{Stage outcome} & \textbf{Representative findings} & \textbf{Cumulative pass} \\
\midrule
Annotator cross-verification (AV-specific checks)
& 240
& 151 required no targeted repair; 89 entered repair for 105 task-level issues.
& Insufficient audio dependence; clue intervals omitting required speech or sound; audio-visual target mismatch; SG boxes following the wrong person; unsupported QA answers.
& \textbf{151/240} (\textbf{62.9\%}) \\
\midrule
Annotator revision and lead-expert acceptance
& 89
& 24 TG, 70 SG, and 11 QA annotations corrected; all 89 videos accepted.
& TG intervals starting too early; SG references, questions, and boxes identifying different targets; QA type or evidence mismatch; imbalanced distractors.
& \textbf{240/240} (\textbf{100\%}) \\
\midrule
Final release audit
& 240
& All AV cases are included only after lead-expert acceptance and audit.
& Required native-audio evidence; aligned audio-visual clues; the same target and answer across settings; valid TG, SG, and QA annotations.
& \textbf{240/240} (\textbf{100\%}) \\
\bottomrule
\end{tabular}
\end{table*}

\begin{table*}[t]
\centering
\caption{
\textbf{Estimated human effort for OmniCompass construction and quality assurance.}
Hours are reconstructed from dated project records using an eight-hour workday. Automated processing, inference, and waiting time are excluded.
}
\label{tab:annotation_effort}
\footnotesize
\setlength{\tabcolsep}{4pt}
\renewcommand{\arraystretch}{1.12}
\begin{tabular}{p{0.16\textwidth} p{0.13\textwidth} p{0.43\textwidth} p{0.12\textwidth} p{0.08\textwidth}}
\toprule
\textbf{Stage} & \textbf{Personnel} & \textbf{Scope} & \textbf{Person-days} & \textbf{Hours} \\
\midrule
Guideline training
& 4 annotators
& Two-day guideline training and pilot-protocol calibration.
& 8
& 64 \\

Initial task construction
& 4 annotators
& Paired \textit{Text} and \textit{Text+Image} TG, SG, and QA instances for 820 VL videos, including the 240-video AV subset.
& 112
& 896 \\

Mutual cross-verification and targeted revision
& 4 annotators
& Mutual cross-verification and corrections within the 354 flagged VL cases and track-specific AV repairs identified during verification.
& 40
& 320 \\

Final acceptance and correction
& Lead expert
& Final acceptance of annotator-confirmed cases, with remaining corrections for target pairing, reference necessity, TG/SG validity, QA evidence, and AV dependence.
& 31
& 248 \\

Repair adjudication
& Lead expert
& Repair-package review, disputed-case resolution, and re-verification of corrected annotations.
& 16
& 128 \\

Lead-expert final release audit
& Lead expert
& Paired-setting consistency, dashboard readback, residual-error checks, and final release validation.
& 16
& 128 \\
\midrule
\textbf{Total}
& \textbf{4 annotators + 1 lead expert}
& \textbf{End-to-end human construction and quality assurance.}
& \textbf{223}
& \textbf{1,784} \\
\bottomrule
\end{tabular}
\end{table*}

\endgroup

\clearpage
\section{Evaluation Protocols and Prompts}
\label{sec:appendix_eval_details}

\subsection{Inputs and Paired Settings}
Frames are sampled uniformly across full videos.
The standard budget is 64 for proprietary models and open-source models above 100B total parameters, and 32 otherwise.
GPT-series runs use 49 frames to reserve a reference slot within the evaluated 50-image input limit; VITA-1.5-7B uses 16.
Budgets remain fixed within a model across tasks and reference settings.
Cross-model results therefore reflect the stated input budgets, not a strictly equal-compute comparison.

For the 52-second VL clip in Figure~\ref{fig:eval_prompt_tg}, the 64-frame inputs are:
\begin{center}
\small
\renewcommand{\arraystretch}{1.12}
\begin{tabular}{@{}ll@{}}
\textit{Text:} & \texttt{Timestamp: 0 seconds. <frame 0>} $\rightarrow$\\
& \texttt{Timestamp: 0.84 seconds. <frame 1>} $\rightarrow$\\
& \texttt{...}\\
& \texttt{Timestamp: 51.96 seconds. <frame 63>} $\rightarrow$\\
& \texttt{[Prompt]}\\[5pt]
\textit{Text+Image:} & \texttt{Timestamp: 0 seconds. <frame 0>} $\rightarrow$\\
& \texttt{Timestamp: 0.84 seconds. <frame 1>} $\rightarrow$\\
& \texttt{...}\\
& \texttt{Timestamp: 51.96 seconds. <frame 63>} $\rightarrow$\\
& \texttt{<ref image>} $\rightarrow$ \texttt{[Prompt]}
\end{tabular}
\end{center}

For the 270-second AV clip in Figure~\ref{fig:eval_prompt_av_tg}, audio follows the sampled frames:
\begin{center}
\small
\renewcommand{\arraystretch}{1.12}
\begin{tabular}{@{}ll@{}}
\textit{Text:} & \texttt{Timestamp: 0 seconds. <frame 0>} $\rightarrow$\\
& \texttt{Timestamp: 4.3 seconds. <frame 1>} $\rightarrow$\\
& \texttt{...}\\
& \texttt{Timestamp: 269.967 seconds. <frame 63>} $\rightarrow$\\
& \texttt{<audio>} $\rightarrow$ \texttt{[Prompt]}\\[5pt]
\textit{Text+Image:} & \texttt{Timestamp: 0 seconds. <frame 0>} $\rightarrow$\\
& \texttt{Timestamp: 4.3 seconds. <frame 1>} $\rightarrow$\\
& \texttt{...}\\
& \texttt{Timestamp: 269.967 seconds. <frame 63>} $\rightarrow$\\
& \texttt{<audio>} $\rightarrow$ \texttt{<ref image>} $\rightarrow$ \texttt{[Prompt]}
\end{tabular}
\end{center}

Each timestamp is literal text in seconds from the start of the video, placed immediately before its corresponding frame, following the design of TimeLens~\citep{zhang2026timelens}.
The frame numbers index sampled frames from zero: \texttt{<frame 0>} is the first, and \texttt{<frame 63>} is the last in these 64-frame examples.
Angle-bracketed items denote media content, not literal text tokens.
\texttt{<audio>} is the video's audio track, and \texttt{<ref image>} is the task-specific reference image.
The arrows indicate input order, and \texttt{...} omits intermediate timestamp--frame pairs; neither is sent to the model.
\texttt{[Prompt]} stands for the task prompt, which includes the full video duration.
The timestamps shown are specific to these clips, not fixed for other videos or frame budgets.
Within each pair, the video evidence, task semantics, and ground truth are shared; only the target specification changes.

TG returns temporal boundaries, and QA selects one of five answer options.
Following V-STaR~\citep{cheng2025vstar}, SG receives the relevant interval and returns one bounding box at each endpoint.
It therefore evaluates target localization at supplied times, not temporal search or dense tracking throughout the interval.
Figures~\ref{fig:eval_prompt_tg}--\ref{fig:eval_prompt_av_qa} illustrate response formats with annotated ground truth.
The output panels are never supplied to the models in either setting.

\subsection{Prompt Examples}
Figures~\ref{fig:eval_prompt_tg}--\ref{fig:eval_prompt_qa} and~\ref{fig:eval_prompt_av_tg}--\ref{fig:eval_prompt_av_qa} show OmniCompass-VL and OmniCompass-AV prompts, respectively.

\begingroup
\makeatletter
\setlength{\@fptop}{0pt}
\setlength{\@dblfptop}{0pt}
\makeatother
\clearpage
\clearpage
\begin{figure*}[t]
\centering
\begin{minipage}[t]{0.485\textwidth}
\begin{evalpromptcard}[\textit{Text}]{13.8cm}
\footnotesize
\evalsectionlabel{Input}
\evalvideoinput
\evalinputpromptdivider
\evalsectionlabel{Prompt}
\begin{evalpromptbody}{5.8cm}
\evalpromptintrovideo

\smallskip
\textbf{Question:} When in the video did Team USA player No.~7 score her first basket?

\smallskip
\textbf{Output format:} Answer with only this JSON object, where
\texttt{start\_time} and \texttt{end\_time} are measured in seconds from
the beginning of the video:

\smallskip
\texttt{\{"start\_time": <number>,}\\
\texttt{"end\_time": <number>\}}
\end{evalpromptbody}
\evalpromptoutput{\scriptsize
\texttt{\{"start\_time": 48.0,}\\
\texttt{"end\_time": 50.0\}}}
\end{evalpromptcard}
\end{minipage}\hfill
\begin{minipage}[t]{0.485\textwidth}
\begin{evalpromptcard}[\textit{Text+Image}]{13.8cm}
\footnotesize
\evalsectionlabel{Input}
\evalvideoreferenceinput{\includegraphics[height=1.72cm,width=0.28\linewidth,keepaspectratio]{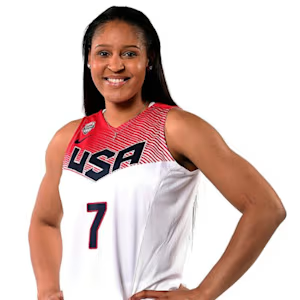}}
\evalinputpromptdivider
\evalsectionlabel{Prompt}
\begin{evalpromptbody}{5.8cm}
\evalpromptintroref

\smallskip
\textbf{Question:} When in the video did the player in the reference image score her first basket?

\smallskip
\textbf{Output format:} Answer with only this JSON object, where
\texttt{start\_time} and \texttt{end\_time} are measured in seconds from
the beginning of the video:

\smallskip
\texttt{\{"start\_time": <number>,}\\
\texttt{"end\_time": <number>\}}
\end{evalpromptbody}
\evalpromptoutput{\scriptsize
\texttt{\{"start\_time": 48.0,}\\
\texttt{"end\_time": 50.0\}}}
\end{evalpromptcard}
\end{minipage}
\caption{\textbf{Temporal Grounding prompts for OmniCompass-VL.}
Both settings receive the same uniformly sampled, timestamped video frames without audio and predict the same event interval. Only the target
specification changes from a textual description to a reference image.}
\label{fig:eval_prompt_tg}
\end{figure*}

\clearpage
\clearpage
\begin{figure*}[t]
\centering

\begin{minipage}[t]{0.485\textwidth}
\begin{evalpromptcard}[\textit{Text}]{18.4cm}
\footnotesize

\evalsectionlabel{Input}
\evalvideoinput
\evalinputpromptdivider

\evalsectionlabel{Prompt}
\begin{evalpromptbody}{10.0cm}
\evalpromptintrovideo

\smallskip
\textbf{Question:} Where in the frame is Team USA player No.~7 from when
she begins her shooting motion until the ball goes in?

\smallskip
The relevant interval starts at 48.72 seconds and ends at 50.12 seconds.
Locate the target described in the question at each of these two timestamps
and report one bounding box for each frame.

Write each box in \texttt{[x1, y1, x2, y2]} order, corresponding to
\texttt{[left, top, right, bottom]}. Normalize all coordinates from 0 to
1000, with the top-left of the frame as the origin.

\smallskip
\textbf{Output format:} Return only this JSON object:

\smallskip
\texttt{\{"start\_time": 48.72,}\\
\texttt{"start\_bbox": [\textless x1\textgreater, \textless y1\textgreater,
\textless x2\textgreater, \textless y2\textgreater],}\\
\texttt{"end\_time": 50.12,}\\
\texttt{"end\_bbox": [\textless x1\textgreater, \textless y1\textgreater,
\textless x2\textgreater, \textless y2\textgreater]\}}
\end{evalpromptbody}

\evalpromptoutput{\scriptsize
\texttt{\{"start\_time": 48.72,}\\
\texttt{"start\_bbox": [587.61, 473.21, 628.53, 666.34],}\\
\texttt{"end\_time": 50.12,}\\
\texttt{"end\_bbox": [644.16, 536.71, 694.75, 833.00]\}}}

\end{evalpromptcard}
\end{minipage}
\hfill
\begin{minipage}[t]{0.485\textwidth}
\begin{evalpromptcard}[\textit{Text+Image}]{18.4cm}
\footnotesize

\evalsectionlabel{Input}
\evalvideoreferenceinput{\includegraphics[height=1.72cm,width=0.28\linewidth,keepaspectratio]{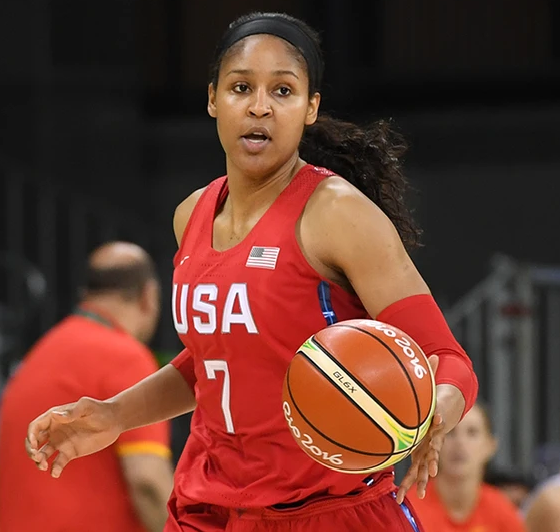}}
\evalinputpromptdivider

\evalsectionlabel{Prompt}
\begin{evalpromptbody}{10.0cm}
\evalpromptintroref

\smallskip
\textbf{Question:} Where in the frame is the player in the reference image
from when she begins her shooting motion until the ball goes in?

\smallskip
The relevant interval starts at 48.72 seconds and ends at 50.12 seconds.
Locate the target described in the question at each of these two timestamps
and report one bounding box for each frame.

Write each box in \texttt{[x1, y1, x2, y2]} order, corresponding to
\texttt{[left, top, right, bottom]}. Normalize all coordinates from 0 to
1000, with the top-left of the frame as the origin.

\smallskip
\textbf{Output format:} Return only this JSON object:

\smallskip
\texttt{\{"start\_time": 48.72,}\\
\texttt{"start\_bbox": [\textless x1\textgreater, \textless y1\textgreater,
\textless x2\textgreater, \textless y2\textgreater],}\\
\texttt{"end\_time": 50.12,}\\
\texttt{"end\_bbox": [\textless x1\textgreater, \textless y1\textgreater,
\textless x2\textgreater, \textless y2\textgreater]\}}
\end{evalpromptbody}

\evalpromptoutput{\scriptsize
\texttt{\{"start\_time": 48.72,}\\
\texttt{"start\_bbox": [587.61, 473.21, 628.53, 666.34],}\\
\texttt{"end\_time": 50.12,}\\
\texttt{"end\_bbox": [644.16, 536.71, 694.75, 833.00]\}}}

\end{evalpromptcard}
\end{minipage}

\caption{\textbf{Spatial Grounding prompts for OmniCompass-VL.}
Both settings receive the same uniformly sampled, timestamped video frames and the same temporal interval. They localize the same target at the two
interval endpoints; only the target specification changes.}
\label{fig:eval_prompt_sg}
\end{figure*}

\clearpage
\clearpage
\begin{figure*}[t]
\centering
\begin{minipage}[t]{0.485\textwidth}
\begin{evalpromptcard}[\textit{Text}]{13.6cm}
\footnotesize
\evalsectionlabel{Input}
\evalvideoinput
\evalinputpromptdivider
\evalsectionlabel{Prompt}
\begin{evalpromptbody}{6.5cm}
\evalpromptintrovideo

\smallskip
\textbf{Question:} How many points does Team USA player No.~7 score in total in the video?

\smallskip
\begin{tabular}{@{}ll@{}}
A. & 0 points\\
B. & 2 points\\
C. & 3 points\\
D. & 4 points\\
E. & 1 point
\end{tabular}

\smallskip
You must choose the most appropriate answer from the five options
(A, B, C, D, or E). If uncertain, make your best choice based on the
provided inputs. Output only the option letter.
\end{evalpromptbody}
\evalpromptoutput{\texttt{B}}
\end{evalpromptcard}
\end{minipage}\hfill
\begin{minipage}[t]{0.485\textwidth}
\begin{evalpromptcard}[\textit{Text+Image}]{13.6cm}
\footnotesize
\evalsectionlabel{Input}
\evalvideoreferenceinput{\includegraphics[height=1.72cm,width=0.28\linewidth,keepaspectratio]{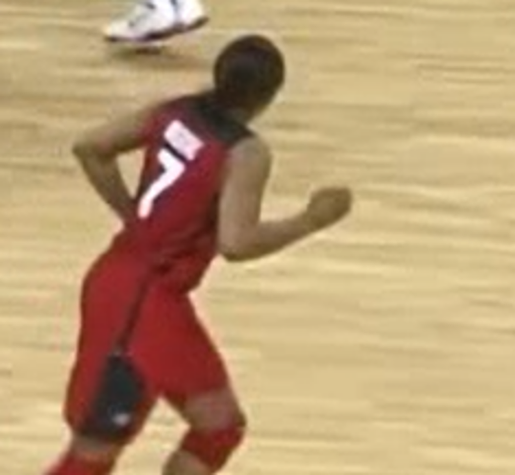}}
\evalinputpromptdivider
\evalsectionlabel{Prompt}
\begin{evalpromptbody}{6.5cm}
\evalpromptintroref

\smallskip
\textbf{Question:} How many points does the player in the reference image score in total in the video?

\smallskip
\begin{tabular}{@{}ll@{}}
A. & 0 points\\
B. & 2 points\\
C. & 3 points\\
D. & 4 points\\
E. & 1 point
\end{tabular}

\smallskip
You must choose the most appropriate answer from the five options
(A, B, C, D, or E). If uncertain, make your best choice based on the
provided inputs. Output only the option letter.
\end{evalpromptbody}
\evalpromptoutput{\texttt{B}}
\end{evalpromptcard}
\end{minipage}
\caption{\textbf{Question Answering prompts for OmniCompass-VL.}
Both settings receive the same uniformly sampled, timestamped video frames and retain the same answer choices and ground truth. Only the target
specification changes.}
\label{fig:eval_prompt_qa}
\end{figure*}

\clearpage
\clearpage
\begin{figure*}[t]
\centering
\begin{minipage}[t]{0.485\textwidth}
\begin{evalpromptcard}[\textit{Text}]{15.2cm}
\footnotesize
\evalsectionlabel{Input}
\evalavvideoinput
\evalinputpromptdivider
\evalsectionlabel{Prompt}
\begin{evalpromptbody}{7.2cm}
\evalpromptintroavvideo

\smallskip
\textbf{Question:} When in the video does the female news anchor with long
brown curly hair, a black-and-white houndstooth jacket, and a black turtleneck
report that France is seeking to redefine its role in Africa?

\smallskip
\textbf{Output format:} Answer with only this JSON object, where
\texttt{start\_time} and \texttt{end\_time} are measured in seconds from
the beginning of the video:

\smallskip
\texttt{\{"start\_time": <number>,}\\
\texttt{"end\_time": <number>\}}
\end{evalpromptbody}
\evalpromptoutput{\scriptsize
\texttt{\{"start\_time": 89.0,}\\
\texttt{"end\_time": 92.0\}}}
\end{evalpromptcard}
\end{minipage}\hfill
\begin{minipage}[t]{0.485\textwidth}
\begin{evalpromptcard}[\textit{Text+Image}]{15.2cm}
\footnotesize
\evalsectionlabel{Input}
\evalavreferencevideoinput{\includegraphics[height=1.72cm,width=0.28\linewidth,keepaspectratio]{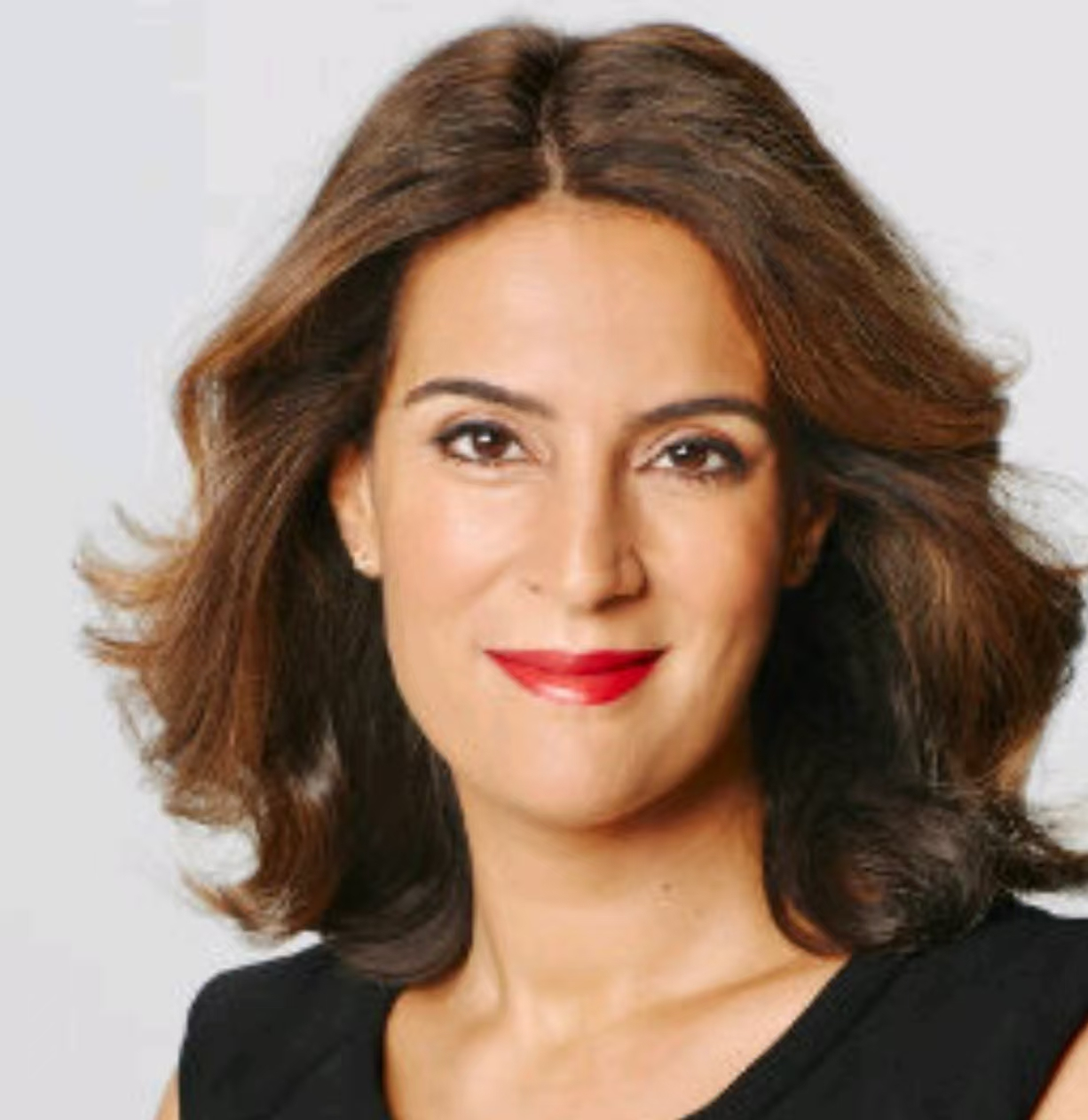}}
\evalinputpromptdivider
\evalsectionlabel{Prompt}
\begin{evalpromptbody}{7.2cm}
\evalpromptintroavref

\smallskip
\textbf{Question:} When in the video does the news anchor in the reference
image report that France is seeking to redefine its role in Africa?

\smallskip
\textbf{Output format:} Answer with only this JSON object, where
\texttt{start\_time} and \texttt{end\_time} are measured in seconds from
the beginning of the video:

\smallskip
\texttt{\{"start\_time": <number>,}\\
\texttt{"end\_time": <number>\}}
\end{evalpromptbody}
\evalpromptoutput{\scriptsize
\texttt{\{"start\_time": 89.0,}\\
\texttt{"end\_time": 92.0\}}}
\end{evalpromptcard}
\end{minipage}
\caption{\textbf{Temporal Grounding prompts for OmniCompass-AV.}
Both settings receive the same timestamped video frames and corresponding
audio, and predict the same audio-visual event interval. Only the target
specification changes.}
\label{fig:eval_prompt_av_tg}
\end{figure*}

\clearpage
\clearpage
\begin{figure*}[t]
\centering
\begin{minipage}[t]{0.485\textwidth}
\begin{evalpromptcard}[\textit{Text}]{20.1cm}
\footnotesize
\evalsectionlabel{Input}
\evalavvideoinput
\evalinputpromptdivider
\evalsectionlabel{Prompt}
\begin{evalpromptbody}{11.4cm}
\evalpromptintroavvideo

\medskip
\textbf{Question:} Where in the frame is the female news anchor with long
brown curly hair, a black-and-white houndstooth jacket, and a black turtleneck
when she reports that French President Emmanuel Macron has arrived in Nairobi,
Kenya?

\medskip
The relevant interval starts at 85.03 seconds and ends at 87.87 seconds.
Locate the target described in the question at each of these two timestamps
and report one bounding box for each frame.

Write each box in \texttt{[x1, y1, x2, y2]} order, corresponding to
\texttt{[left, top, right, bottom]}. Normalize all coordinates from 0 to 1000,
with the top-left of the frame as the origin.

\medskip
\textbf{Output format:} Return only this JSON object:

\smallskip
\texttt{\{"start\_time": 85.03,}\\
\texttt{"start\_bbox": [<x1>, <y1>, <x2>, <y2>],}\\
\texttt{"end\_time": 87.87,}\\
\texttt{"end\_bbox": [<x1>, <y1>, <x2>, <y2>]\}}
\end{evalpromptbody}
\evalpromptoutput{\scriptsize
\texttt{\{"start\_time": 85.03,}\\
\texttt{"start\_bbox": [278.96, 63.39, 749.11, 755.36],}\\
\texttt{"end\_time": 87.87,}\\
\texttt{"end\_bbox": [276.17, 68.35, 735.16, 760.32]\}}}
\end{evalpromptcard}
\end{minipage}\hfill
\begin{minipage}[t]{0.485\textwidth}
\begin{evalpromptcard}[\textit{Text+Image}]{20.1cm}
\footnotesize
\evalsectionlabel{Input}
\evalavreferencevideoinput{\includegraphics[height=1.72cm,width=0.28\linewidth,keepaspectratio]{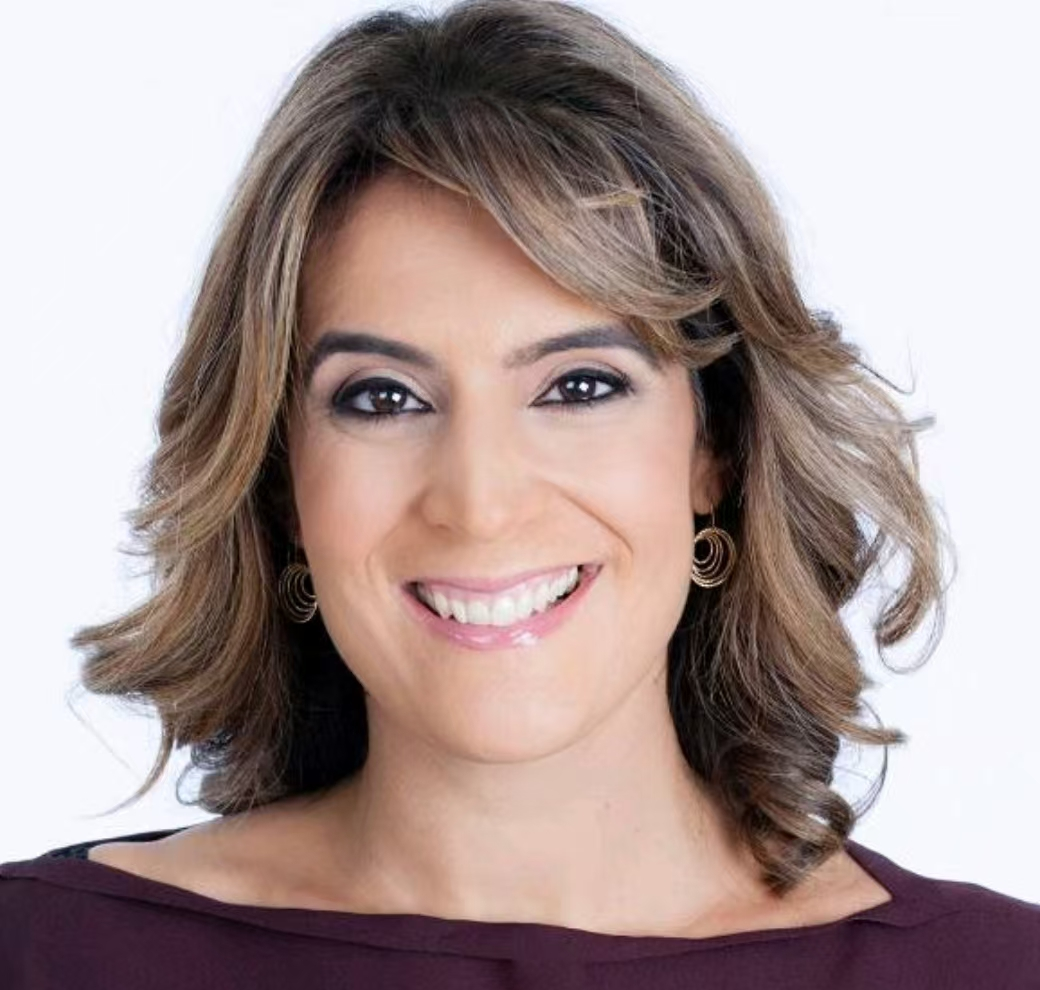}}
\evalinputpromptdivider
\evalsectionlabel{Prompt}
\begin{evalpromptbody}{11.4cm}
\evalpromptintroavref

\medskip
\textbf{Question:} Where in the frame is the news anchor in the reference
image when she reports that French President Emmanuel Macron has arrived in
Nairobi, Kenya?

\medskip
The relevant interval starts at 85.03 seconds and ends at 87.87 seconds.
Locate the target described in the question at each of these two timestamps
and report one bounding box for each frame.

Write each box in \texttt{[x1, y1, x2, y2]} order, corresponding to
\texttt{[left, top, right, bottom]}. Normalize all coordinates from 0 to 1000,
with the top-left of the frame as the origin.

\medskip
\textbf{Output format:} Return only this JSON object:

\smallskip
\texttt{\{"start\_time": 85.03,}\\
\texttt{"start\_bbox": [<x1>, <y1>, <x2>, <y2>],}\\
\texttt{"end\_time": 87.87,}\\
\texttt{"end\_bbox": [<x1>, <y1>, <x2>, <y2>]\}}
\end{evalpromptbody}
\evalpromptoutput{\scriptsize
\texttt{\{"start\_time": 85.03,}\\
\texttt{"start\_bbox": [278.96, 63.39, 749.11, 755.36],}\\
\texttt{"end\_time": 87.87,}\\
\texttt{"end\_bbox": [276.17, 68.35, 735.16, 760.32]\}}}
\end{evalpromptcard}
\end{minipage}
\caption{\textbf{Spatial Grounding prompts for OmniCompass-AV.}
Both settings receive the same timestamped video frames, corresponding audio,
and temporal interval. They localize the same target at the two interval
endpoints; only the target specification changes.}
\label{fig:eval_prompt_av_sg}
\end{figure*}

\clearpage
\clearpage
\begin{figure*}[t]
\centering
\begin{minipage}[t]{0.485\textwidth}
\begin{evalpromptcard}[\textit{Text}]{16.8cm}
\footnotesize
\evalsectionlabel{Input}
\evalavvideoinput
\evalinputpromptdivider
\evalsectionlabel{Prompt}
\begin{evalpromptbody}{9.5cm}
\evalpromptintroavvideo

\smallskip
\textbf{Question:} What happens at the end of the preceding report, just
before the female news anchor with curly brown hair and a black-and-white
patterned jacket reports that thousands of fans have gathered in Vienna?

\smallskip
\begingroup
\setlength{\tabcolsep}{2pt}
\renewcommand{\arraystretch}{0.95}
\begin{tabular}{@{}lp{0.86\linewidth}@{}}
A. & Passengers disembark at a quay\\
B. & A white-haired woman is embraced at an awards ceremony\\
C. & Flag-waving fans crowd Vienna's streets\\
D. & Two studio presenters discuss a political crisis\\
E. & An evacuation plane waits on the apron
\end{tabular}
\endgroup

\smallskip
You must choose the most appropriate answer from the five options
(A, B, C, D, or E). If uncertain, make your best choice based on the
provided inputs. Output only the option letter.
\end{evalpromptbody}
\evalpromptoutput{\texttt{B}}
\end{evalpromptcard}
\end{minipage}\hfill
\begin{minipage}[t]{0.485\textwidth}
\begin{evalpromptcard}[\textit{Text+Image}]{16.8cm}
\footnotesize
\evalsectionlabel{Input}
\evalavreferencevideoinput{\includegraphics[height=1.72cm,width=0.28\linewidth,keepaspectratio]{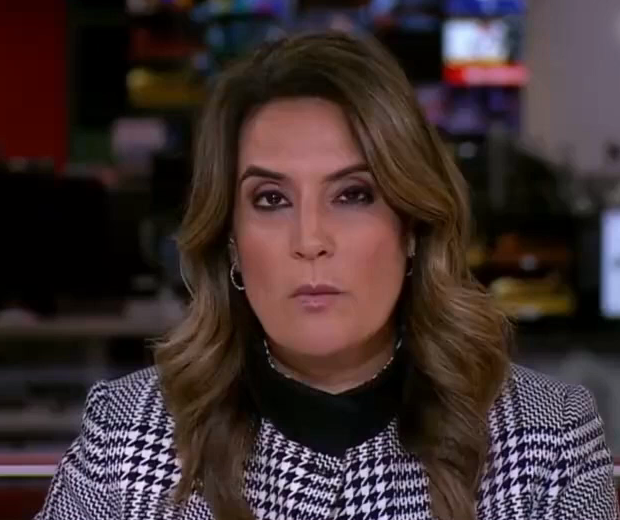}}
\evalinputpromptdivider
\evalsectionlabel{Prompt}
\begin{evalpromptbody}{9.5cm}
\evalpromptintroavref

\smallskip
\textbf{Question:} What happens at the end of the preceding report, just
before the news anchor in the reference image reports that thousands of fans
have gathered in Vienna?

\smallskip
\begingroup
\setlength{\tabcolsep}{2pt}
\renewcommand{\arraystretch}{0.95}
\begin{tabular}{@{}lp{0.86\linewidth}@{}}
A. & Passengers disembark at a quay\\
B. & A white-haired woman is embraced at an awards ceremony\\
C. & Flag-waving fans crowd Vienna's streets\\
D. & Two studio presenters discuss a political crisis\\
E. & An evacuation plane waits on the apron
\end{tabular}
\endgroup

\smallskip
You must choose the most appropriate answer from the five options
(A, B, C, D, or E). If uncertain, make your best choice based on the
provided inputs. Output only the option letter.
\end{evalpromptbody}
\evalpromptoutput{\texttt{B}}
\end{evalpromptcard}
\end{minipage}
\caption{\textbf{Question Answering prompts for OmniCompass-AV.}
Both settings receive the same timestamped video frames, corresponding audio,
and answer candidates. The correct answer is unchanged across target
specifications.}
\label{fig:eval_prompt_av_qa}
\end{figure*}

\endgroup
\clearpage
\endgroup

\end{document}